\documentclass[10pt, conference, compsocconf]{IEEEtran}
\IEEEoverridecommandlockouts
\ifCLASSINFOpdf
  \usepackage[pdftex]{graphicx}
\else
\fi
\graphicspath{{./pics/}}

\usepackage{array}
\usepackage{amsmath}
\usepackage{color}
\usepackage{url}
\usepackage{adjustbox}
\usepackage{multirow}
\usepackage{algorithmic}
\usepackage{algorithm}
\usepackage{hyperref}

\newcolumntype{C}[1]{>{\centering\let\newline\\\arraybackslash\hspace{0pt}}m{#1}}

\usepackage{amsfonts}
\usepackage{multirow}
\usepackage{caption} 
\usepackage{subfig}
\usepackage{graphics}

\begin{document}
\nocite{*}

%
\title{OF-AE: Oblique Forest AutoEncoders\thanks{This work was supported by a grant of the Romanian Ministry of Research and Innovation, CCCDI – UEFISCDI, project number 178PCE/2021, PN-III-P4-ID-PCE-2020-0788, \textit{Object PErception and Reconstruction with deep neural Architectures} (\textit{OPERA}), within PNCDI III.}}


%
\author{\IEEEauthorblockN{Cristian Daniel Alecsa\IEEEauthorrefmark{2}}
\IEEEauthorblockA{\IEEEauthorrefmark{2}Romanian Institute of Science and Technology, Romania\\ \IEEEauthorrefmark{2}Technical University of Cluj-Napoca, Romania \\
Email: alecsa@rist.ro}
}


\maketitle

\begin{abstract}
In the present work we propose an unsupervised ensemble method consisting of oblique trees that can address the task of auto-encoding, namely Oblique Forest AutoEncoders (briefly \textbf{\textit{OF-AE}}). Our method is a natural extension of the \emph{eForest} encoder introduced in \cite{eForest}. More precisely, by employing oblique splits consisting in multivariate linear combination of features instead of the axis-parallel ones, we will devise an auto-encoder method through the computation of a sparse solution of a set of linear inequalities consisting of feature values constraints. The code for reproducing our results is available at \url{https://github.com/CDAlecsa/Oblique-Forest-AutoEncoders}.
\end{abstract}




%
\IEEEpeerreviewmaketitle

\section{Introduction}
\label{sec:intro}
It is well known that \textit{Classification and Regression Trees} (briefly CART) \cite{CART} have proven to be very successful methods for various data analysis problems. The original CART algorithm partitions the feature space using axis-parallel splits. The training of a classical decision tree $\mathcal{T}$ relies on $\emph{greedy optimization}$, i.e. the root of the tree is the whole input space $\mathcal{X}$ which is split into two disjoint regions, and this process continues in a recursive manner. More precisely, when a sample $x \in \mathcal{X}$ reaches a decision node it will be sent left or right based upon a binary decision function of the form $x_j > z$ where $j$ is a feature and $z$ is the threshold (the position of the cut along the $j^{th}$ coordinate), where both the feature and the threshold are determined during training. Even though decision trees are inherently interpretable and are very fast from a computational point of view, there are numerous extensions to classical CART methods. The extremely randomized trees (briefly ERT) \cite{GeurtsERT} induces randomization into the optimization of the trees by selecting the threshold fully at random for each candidate feature. The ERT algorithm is implemented in \texttt{SKLearn} as \textit{ExtraTreeRegressor} and for many benchmark regression and classification problems, it achieves similar results as the classical decision trees. \\

A popular extension of CART is the \emph{Random Forest} algorithm, briefly RF, introduced by Leo Breiman in \cite{RF} (see also \cite{BaggingPredictors}). The RF method consists of different randomized decision trees $\mathcal{T}_1, \ldots, \mathcal{T}_N$ which are trained independently and which are finally aggregated all together with the average of the underlying individual scores. The randomness induced by the RF method which helps reducing the correlation between each individual tree, consists of training each decision tree on a randomly selected subset of the training data and using a random subset of features in each node of every tree. \\

Even though CART and RF are suited for prediction tasks, they can also be perceived as clustering methods. In \cite{LinRF}, Lin and Jeon presented an interesting connection between RF and nearest neighbor predictors. At the same time, Moosmann et. al. \cite{ERCForest} introduced the Extremely Randomized Clustering Forest (briefly ERCForest) which is an ensemble of clustering trees, based on spatial partitioning that assigns a distinct region label to each of the leafs (see also \cite{ClusteringRF}). The usage of the clustering method of the ERCForest can be observed in the unsupervised algorithm \emph{RandomTreesEmbedding} from \texttt{SKLearn}, where the data points are clustered according to which leaf they fall in. Furthermore, it is worth noticing that the ERCForest is eventually related to Clustering Trees (CT) introduced in \cite{ClusteringDT} that are Decision Trees able to find natural clusters in very high dimensional spaces. \\

A different extension to classical CART methods represent the so-called \emph{Soft Decision Trees} (SDT) considered in \cite{IrsoySFDT} and \cite{FrosstHinton} which can be considered as the probabilistic variants of CART. More exactly, these methods are based upon the idea of splitting the parameters of the tree via gradient descent-type methods, by relaxing the hard decision functions to soft probabilistic decisions. A similar idea whas developed for the so-called \emph{Probabilistic Random Forest} (PRF) introduced in \cite{PRF} where the underlying algorithm takes into account uncertainty in the measurements of the features and of the assigned labels. In PRF, the uncertainty in the features is represented through the distribution function of each feature for every sample point, while in SDT \cite{IrsoySFDT} the decision function (also called gating function) is represented by the sigmoid mapping applied to a linear combination of features. An extension of these types of decision trees represents \emph{Deep Neural Decision Forest} (DNDF) introduced in \cite{DNDF} where each node is represented by a neural network layer and where the backpropagation algorithm is used in the training process. An extension of DNDF are the \emph{Adaptive Neural Trees} (ANT) from the work \cite{TannoANT}, where each internal node contains a router (decision) function represented by a neural network, every edge is represented by a neural layer through a so-called transformer, and also every leaf node contains a solver which maps the transformed input data and estimates the conditional distribution of the labels. We highlight that these types of trees are optimized with respect to a global loss function through gradient-based methods. The nodes in the DNDF and ANT uses the same input $x$ belonging to the dataset, unlike the classical neural networks where the output of the previous layer is the input of the next layer. At the same time, SDT and its variants rely on hierarchical decisions, while neural networks are based upon hierarchical features. Finally, we mention that, for the CART algorithm, the hard thresholding imposes a certain clustering structure, i.e. an object reaches a leaf node and the prediction value is given with respect to the sample points in that node. On the other hand, in the case of probabilistic routing, each object reaches all the leaf nodes with some probability and for the final prediction one must take into consideration the weighted predictions given by all leaf nodes. Therefore, probabilistic-type or soft trees do not impose the same natural clustering order as in the case of the CART methods. This can be easily observed in the case of DNDF or ANT where there doesn't exist a "splitting" procedure as in CART, since each batch of inputs goes through all the neural network modules, the latter ones being update based on that sample batch.

\section{Related work}
\label{sec:relwork}

An improvement over the CART algorithms are the \emph{Oblique Decision Trees} (briefly ODT) which partitions the input space $\mathcal{X}$ using multivariate tests. More precisely, these so-called Multivariate Decision Trees test at each internal node several attributes by utilising linear combinations of features of the form 
$$ \sum\limits_{j = 1}^{p} \theta_j x_j > b, $$
where $p$ are the number of features, and the parameters $(\theta_j, b)$ consists of the weights $\theta_j$ and the bias $b$. It is well known that training ODT is in general associated with a high training run-time cost. In more detail, for a given dataset $\mathcal{D}_n$ consisting of $n$ samples with $p$ features, the computational time of one split score is $\mathcal{O}(pn)$ for axis-parallel splits, thus the optimal split at an internal node can be found with a greedy approach by searching for all possible splits along the feature axes in a relatively fast manner (the computational time can be improved by utilising the Extremly Randomized Trees - ERT). On the other hand, Murthy et. al. \cite{Murthy} have shown that the number of splits computed in order to find the best multivariate split by a complete split is impractical since it is of order $\mathcal{O} \left( 2^p C_n^p \right)$. \\

The pioneering work on oblique trees began with the work of Bremain et. al. \cite{CART} which introduced the so-called \emph{Classification and Regression Trees - Linear Combination}, in a nutshell CART-LC. The search for the best multivariate split at an internal node in the CART-LC algorithm was made using a hill-climb approach. \\

There are multiple extensions of the classical Oblique Tree CART-LC. One of them is the recent CO2 algorithm (\emph{Continuous Optimization of Oblique Splits}) from \cite{CO2}, which focuses on optimizing an objective function through gradient-based methods. Instead of minimizing a discontinuous loss function  which makes the distribution of the data sensitive to various changes in the splitting of parameters, the CO2 uses a continuous upper bound for the loss function, along with some constraints on the norm of the parameters. \\

A different methodology is considered in \cite{HHCART} where the authors introduced the so-called \emph{HHCART} oblique decision method which belongs to rotation-based methods. The underlying approach is that the original space where the dataset belongs to is transformed into a feature space. Then, one finds an axis-parallel split in the feature space which corresponds to a multivariate oblique split in the original space. The reflected training samples are given by $\hat{\mathcal{D}} = \mathcal{D} \mathcal{H}$, where $\mathcal{D}$ represents the original training data and the Householder matrix is given by 
$$ \mathcal{H} = \mathcal{I} - 2uu^T, $$
where 
$$ u = \dfrac{e - d}{\| e - d \|_2}. $$
Here, $d$ represents the dominant eigenvector of the estimated covariance matrix corresponding of a class (suggesting the orientation of that class), and $e$ is the standard basis vector corresponding to a chosen feature. Since the Householder matrix $\mathcal{H}$ is symmetric and orthogonal, a sample point in the transformed space can be mapped with a minimal cost to the original space, due to the fact that $\mathcal{H}\mathcal{H} = \mathcal{I}$. Since the \emph{HHCART} algorithm uses a fast method to construct a new feature space where the axis-parallel splits are found, the searching for multivariate splits can be made very fast in the higher-dimensional original space. \\

The last Oblique Tree we recall is the \emph{RandCART} algorithm (see \cite{RandomRF}) that which is similar to the previously presented method \emph{HHCART} due to the fact that an oblique decision split in the original space corresponds to an axis-parallel split in a feature transformed space. In more detail, for \emph{RandCART}, the training samples $\mathcal{D}$ are converted into a new feature space $\mathcal{D}^\prime$ of the same dimension as $\mathcal{D}$. More precisely, one generates randomly and independently $p^2$ terms $m_{ij} \sim \mathcal{N}(0, 1)$ which defines a square matrix $\mathcal{M}$. After that, $\mathcal{M}$ is decomposed using \texttt{QR decomposition} as $\mathcal{M} = QR$, where $Q$ is orthogonal and $R$ is upper triangular. Similar to \emph{HHCART}, the training samples $\mathcal{D}$ are transformed to a new feature space $\hat{\mathcal{D}}$ such that $\hat{\mathcal{D}} = \mathcal{D} Q$. \\
Finally, it is worth saying that all the Oblique Trees methods can be 
easily extended to a bagging approach as in the RF technique. The comparison of various Oblique Trees (along with their precise algorithmic formulation) and their bagging extensions was thoroughly made in \cite{EODT}. \\

We end the present section by turning our focus to the idea of autoencoders, which dates back to the work of \cite{BourlardAE}. A linear autoencoder can be understood as equivalent to the \emph{PCA} technique. By introducing nonlinearities inside the encoder and decoder parts, these methods can be useful for finding low-dimensional representations of the underlying data. Therefore, an autoencoder is a particular class of models which maps the input space to a latent space and then maps the hidden representations back to the original space. There are two cases which can be distinguished, i.e. when the autoencoder has a narrow bottleneck which leads to undercomplete representations and the case of a large bottleneck which is the situation of an overcomplete representation. There are numerous extensions to the basic idea of autoencoders. More explicitly, there are the denoising autoencoders \cite{DenoisingAE} where one uses inputs with noise and then train the model to reconstruct an uncorrupted version of those inputs. Second of all, there exists the variational-type autoencoders introduced in \cite{VAE} which are the probabilistic version of the classical autoencoder. \emph{VAE} is a generative model with the advantage of creating new input vectors by sampling from a well known distribution. This differs significantly from a basic autoencoder which just computes the embeddings of the inputs. \\

Even though all of the aforementioned autoencoder-type models are, in general, neural networks with various structures, only recently autoencoders based upon Decision Trees gain significant attention from the Machine Learning community. The tree-type autoencoders can be categorized into the following groups. \\
The first one, introduced by Irsoy and Alpaydin in \cite{IrsoyAE} is a SDT-type method where a sigmoid gating function is used at every internal node, and where the final structure is based on stacking two SDTs, an encoder and a decoder. Furthermore, the training phase of this type of autoencoder is not based on backpropagation but a layer-by-layer training. \\
The second model which we shall recall is the interpretable autoencoder introduced in the paper of Aguilar et. al. \cite{AguilarAE} where the model architecture is based on multiple DT, where the $i^{th}$ tree is trained with all of the features except the $i^{th}$ attribute where this attribute is considered the target class. \\
The third autoencoder-type model are the Generative Trees (GT) recently introduced in \cite{GTAdversarialCopycat} where the autoencoder-type model is based on a generative tree and on an decision tree which acts like a GAN-type discriminator, respectively. More precisely, the GT represents an extension of GAN-type neural methods to classical DT. Quite interestingly, there are two versions to train the generative tree, namely the adversarial and the copycat technique, respectively. The advantage of these trees is that they perform well on a vary broad type of experiments, such as missing data imputation, training from generated data and generating data augmentation. \\
The last model that we recall is the \emph{eForest} encoder method introduced in \cite{eForest} which represents the backbone of the present article. The \emph{eForest} method can both encode and decode the input vectors. In more detail, the encoding is done by making the input samples go through all of the RF estimators in order to find the path from each tree where every samples goes to. Then, for each sample which will be decoded, it builds up the estimator spaces which represents the restriction on all the features for that particular sample. Moreover, for the decoding part, the \emph{eForest} encoder gathers up the attributes restrictions from all the subspaces and then uses the Maximum Compatibility Rule (briefly MCR) in order to find the lower and upper values of the features for each given sample. Finally, we mention that the \emph{eForest} encoder is able to reconstruct images like the ones belonging to the benchmark datasets \emph{MNIST} and \emph{CIFAR10} with much higher fidelity than many CNN autoencoders. At the same time, it is computationally inexpensive compared with classical neural network autoencoders.

\section{Proposed Methodology}
\label{sec:method}
In this section we shall present the technique that will aid us extending the \emph{eForest} encoder to ODT. In what follows, we shall employ different types of ODT coupled simultaneously into an unsupervised-type \texttt{Bagging Regressor} which will form an encoder-decoder pair. The main difference between our approach for the ODT and the original \emph{eForest} encoder is that the latter one is based upon the Maximum Compatibility Rule. More precisely, the decoder part of the \emph{eForest} consists in taking, for a given sample, the feature restrictions from the right branches of the trees paths, while the final application of the MCR is related to taking the maximum bound of the feature values from these features subspaces. On the other hand, we will show that for the ODT we do not need to employ the MCR, which is itself a heuristic approach, but we can gather, for a given sample which need to be decoded, all of the feature restrictions from the underlying path (by taking into account also the sign of the branches) and form an optimization problem with constraints. Additionally, for the image reconstruction cases we can add auxilliary constraints such that the feature values (which represent the pixel intensities) belong to the closed interval $[0, 255]$. Moreover, for the case of RGB images, we will employ the same technique as in \emph{eForest} by training a different encoder-decoder pair for every channel. \\

\begin{figure*}
\centering
\includegraphics[width=.85\linewidth]{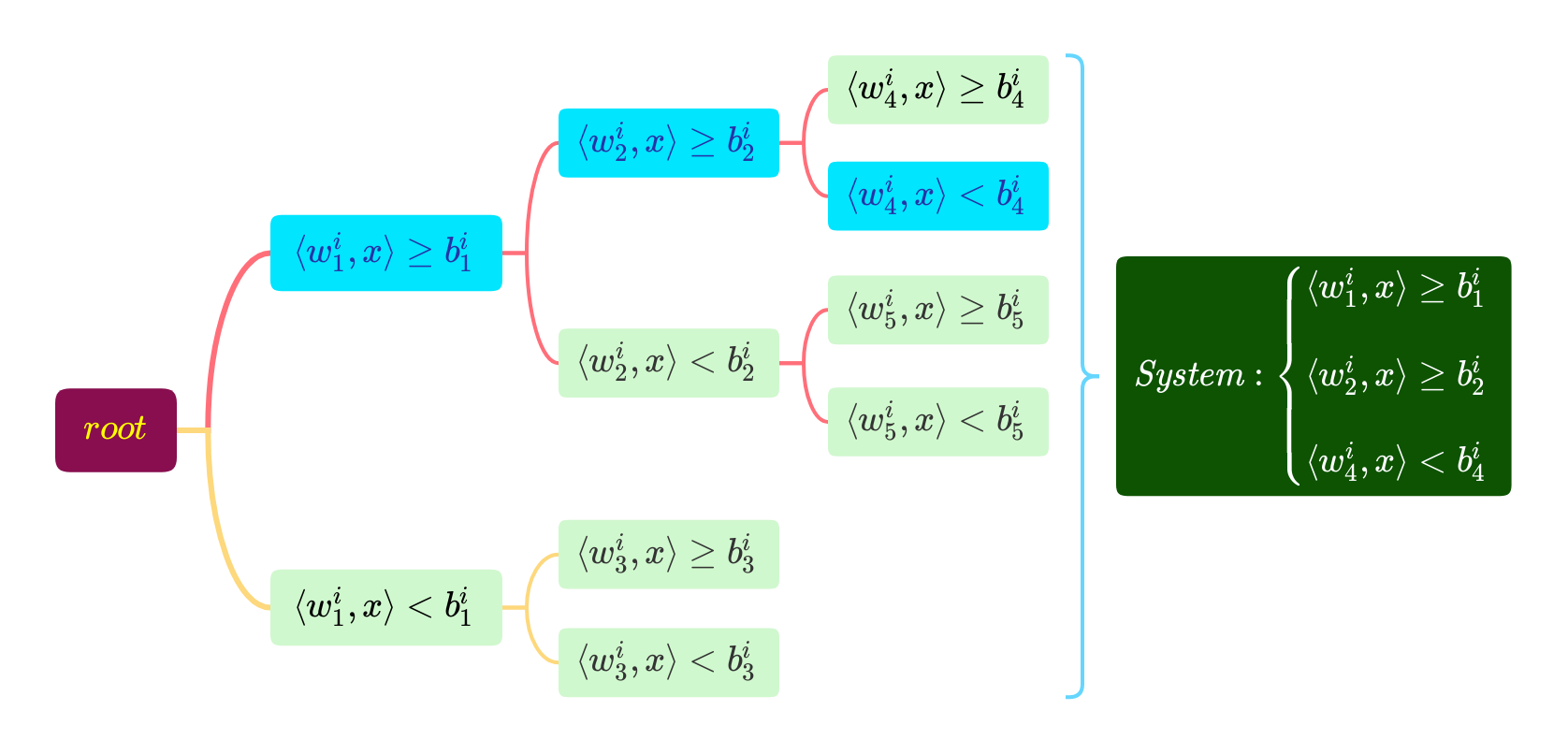}
\caption{A system of constraints obtained from a sample path.}
\label{fig:tree}
\end{figure*}

In what follows we shall propose our encoding-decoding method for a given ensemble of ODT $\mathcal{T}_1, \ldots, \mathcal{T}_N$. For every tree of this ensemble that is already trained, the forward encoding process consists in sending an input vector through each individual tree and to retain the weights and the tresholds for every internal node from the path the samples passes through. For brevity, for some index $i \in \lbrace 1, \ldots, N \rbrace$, let's consider a tree $\mathcal{T}_i$ which is depicted in Figure \eqref{fig:tree} 
Then, for a sample $x \in \mathcal{X}$, the path traversed by the input vector $x$ is highlighted with blue while the other internal nodes not used in the current path are denoted with light green. 

For the tree depicted in \eqref{fig:tree} we observe that the first oblique condition on the root is $\langle w_1^i, x \rangle \geq b_1^i$ for the right branch, and $\langle w_1^i, x \rangle < b_1^i$ for the left branch. Furthermore, for the internal node from the right branch of the sample path, the right branch condition is $\langle w_2^i, x \rangle \geq b_2^i$ while the left branch inequality is $\langle w_2^i, x \rangle < b_2^i$. Finally, for the last internal node of the sample path of $x$, the conditions are $\langle w_4^i, x \rangle \geq b_4^i$ for the right branch and $\langle w_4^i, x \rangle < b_4^i$ for the left branch. Therefore, for the tree $\mathcal{T}_i$ we obtain the following multivariate system of equations based on the weights and the tresholds from the path of $x$:
$$ \left\{\begin{matrix} 
\langle w_1^i, x \rangle \geq b_1^i \\ \\
\langle w_2^i, x \rangle \geq b_2^i \\ \\
\langle -w_4^i, x \rangle > -b_4^i
\end{matrix}\right. $$
From our preliminary simulations we have observed that the final system of equations do not lead to significantly different solutions if we use strict or non-strict inequalities (since we are on the boundary of the weights constraints), hence for the tree depicted in Figure \eqref{fig:tree} we can take
$$ \left\{\begin{matrix} 
\langle w_1^i, x \rangle \geq b_1^i \\ \\
\langle w_2^i, x \rangle \geq b_2^i \\ \\
\langle -w_4^i, x \rangle \geq -b_4^i
\end{matrix}\right. $$

Now, we will formalize our method for the entire ODT ensemble $\mathcal{T}_1, \ldots, \mathcal{T}_N$. For this ensemble, let's consider a generic tree $\mathcal{T}_i$ of a given index $i \in \lbrace 1, \ldots, N \rbrace$. If $\mathcal{T}_i$ has the depth $d_i$ then the maximum number of nodes is $2^{d_i} - 1$. For the root which represents the first node we have the right and left branch conditions $\langle w_1^i, x \rangle \geq b_1^i$ and $\langle -w_1^i, x \rangle \geq -b_1^i$, respectively. For the second node, we obtain the inequalities $\langle w_2^i, x \rangle \geq b_2^i$ and $\langle -w_2^i, x \rangle \geq -b_2^i$. By continuing the argument for the maximum depth, we will have the general form of the right and left inequalities $\langle w_j^i, x \rangle \geq b_j^i$ and $\langle -w_j^i, x \rangle \geq -b_j^i$ for $j \in \lbrace 1, \ldots, n_{i} \rbrace$, where $n_i \leq 2^{d_i} - 1$ is the number of nodes in the ODT $\mathcal{T}_i$. For a sample $x \in \mathcal{X}$ which needs to be encoded, we consider the path of $x$ denoted as $\mathcal{P}_i$ which contains the indices of the nodes in the path, i.e. if $x$ traverses $m_i$ nodes defined by the permutation $\lbrace k_1^i, \ldots, k_{m_i}^i \rbrace$, then we define the path as the set of those node indices, namely $\mathcal{P}_i = \lbrace k_1^i, \ldots, k_{m_i}^i \rbrace$. At the same time, for a node $j \in \mathcal{P}_i$, we will utilize the notation 
$$ sign(j) = 
\begin{cases}
1; & \textit{j goes to the right branch} \\
-1; & \textit{j goes to the left branch}
\end{cases}
$$
By using the above notations, a sample $x$ is encoded through the system of inequalities $\left( sign(j) \langle w^i_j, x \rangle \geq sign(j) b^i_j \right)_{j \in \mathcal{P}_i}$, i.e.
$$ \left\{\begin{matrix} 
sign(k_1^i) \langle w_{k_1}^i, x \rangle \geq sign(k_1^i) b_{k_1}^i \\ \\
sign(k_2^i) \langle w_{k_2}^i, x \rangle \geq sign(k_2^i) b_{k_2}^i \\ \\
\ldots \ldots \ldots \ldots \ldots \ldots \ldots \ldots \ldots \ldots \ldots \\ \\
sign(k_{m_i}^i) \langle w_{k_{m_i}}^i, x \rangle \geq sign(k_{m_i}^i) b_{k_{m_i}}^i
\end{matrix}\right. $$
Now, if we put together all of the sample paths of $x$ from the ensemble trees, we obtain a system of inequalities of the form 
$$ Ax \geq b,$$ 
where 
$$ b = \begin{bmatrix}
sign(k_1^1) b_{k_1}^1 \\
\ldots \\
sign(k_{m_1}^1) b_{k_{m_1}}^1 \\
\ldots \\
sign(k_1^N) b_{k_1}^N \\
\ldots \\ 
sign(k_{m_N}^N) b_{k_{m_N}}^N
\end{bmatrix} \in \mathbb{R}^{m_1 \times m_2 \times \ldots m_N}
$$
and
$$ 
A = \begin{bmatrix}
sign(k_1^1) w_{k_1}^1 \\
\ldots \\
sign(k_{m_1}^1) w_{k_{m_1}}^1 \\
\ldots \\
sign(k_1^N) w_{k_1}^N \\
\ldots \\ 
sign(k_{m_N}^N) w_{k_{m_N}}^N
\end{bmatrix} \in \mathbb{R}^{m_1 \times m_2 \times \ldots m_N \times p},
$$
where $p$ are the number of features. \\

Now, for the decoding process of the sample $x \in \mathcal{X}$ we shall employ the following optimization problem:
\begin{equation*}
\begin{cases}
\textit{minimize } \| x \|_1 \\
\textit{subject to } Ax \geq b,    
\end{cases}
\end{equation*}
where we have chosen the standard $l_1$ norm with the constraints given by the set of linear inequalities $Ax \geq b$. Furthermore, in our codes we made the computing of a sparse solution of the set of the linear inequalities given by the weights and thresholds with the help of the $\texttt{CVXPY}$ package (see the references \cite{CVXPY_1} and \cite{CVXPY_2}, respectively). \\

For our implementations, we have considered only the \emph{HHCART} and the \emph{RandCART} as the choices of ODT. This is due to the fact that these methods are, in general, faster than other types of ODT (like the \emph{CO2}) and they work well on a variety of tasks. At the same time, we have implemented a unsupervised version of the \textit{Bagging Regressor} from \texttt{SKLearn} (\cite{scikit-learn}) by employing another technique from the \texttt{SKLearn}'s implementation of \textit{RandomTreesEmbedding} (which was used in the original version of the \emph{eForest} encoder) where a uniformly one-dimensional target is generated for the tree ensemble. It is worth mentioning that our codes are based upon the implementations from \cite{scikit-obliquetree} and  \url{https://github.com/valevalerio/Ensemble_Of_Oblique_Decision_Trees} related to the thesis \cite{EODT} (where \emph{HHCART} was adapted for regression problems using \emph{MSE} as a criterion). The original \emph{HHCART} implementations use the \emph{PCA} method for transforming the original space to the feature space. But, in our codes, we have used the \emph{PCA} method along with other alternatives (\emph{Truncated-SVD}, \emph{FastICA} and \emph{Gaussian Random Projection}). All these methods are used with the parameter $\textit{n\_components}$ equal to 1, i.e. the feature space is one-dimensional.\\

Finally, in the following sequel, our unsupervised encoding-decoding algorithm constructed with the aid of the $l_1$ constrained optimization problem will be called \textbf{\emph{OF-AE}} which briefly stands for \textbf{Oblique Forest AutoEncoder}.

\section{Experiments}
\label{sec:exp}

In this section we shall present the results for our approach and we will emphasize the tabular and image datasets with which we will work with. Furthermore, we will also consider some interesting remarks about the differences between the classical \emph{eForest} encoder and our ODT approach.

\subsection{Datasets $\&$ Experimental setup}
\label{subsec:data}

To evaluate the performance of our proposed technique, we will consider some benchmark datasets. For tabular datasets, we will consider the following: from \texttt{SKLearn} we will employ the \textit{Diabetes} dataset, and from the \texttt{UCI Machine Learning repository} we shall employ \textit{Seeds} and \textit{HTRU2}, and also \textit{Compas} from
\url{https://github.com/tunguz/TabularBenchmarks/tree/main/datasets}, respectively. On the other hand, for image-type datasets, we will consider as baseline datasets \textit{MNIST} and \textit{CIFAR10} which were already used as benchmarks in the study of the \emph{eForest} encoder from \cite{eForest} (these two classical datasets are downloaded through the \texttt{Keras} API). For image-related datasets, we will also use the \emph{Oxford Flowers} dataset from \url{https://www.robots.ox.ac.uk/~vgg/data/flowers/102/}. \\

Furthermore, we will also utilize a particular unlabeled RGB image dataset used in \cite{SYNASC} (which we will succinctly call it \emph{CHD2R} - \emph{Cultural Heritage Dataset for Digital Reconstruction}) which is comprised of cultural heritage assets (denoted as CH) of all grand museums. We will show the ability of the \textbf{\emph{OF-AE}} to reconstruct the content of textile artefacts with traditional motifs. Along with this, we will show heuristically the capability of our proposed autoencoder to digitally reconstruct archeologically inspired images which contain a large disparity of colors on very small areas. Furthermore, we will also empirically investigate how an \textbf{\emph{OF-AE}} trained on \emph{CIFAR10} can help us decode images from \emph{CHD2R}. \\ 

For RGB-type images, similar to \emph{eForest} encoder, we will consider for each channel a different \textbf{\emph{OF-AE}} model and then stack the results with respect to the three channels in order to fully decode an image. Now, in contrast to the \emph{eForest} we will also investigate the effects of different parameters in order to study the stability of our proposed method. \\

\begin{table}[!h]
\normalsize
\renewcommand{\arraystretch}{1.17}
\begin{tabular}{ ||p{2.4cm}||p{2.4cm}||p{2.4cm}||  }
 \hline
 \multicolumn{3}{||c||}{Datasets} \\
 \hline\hline
 Name& Samples &Features \\
 \hline\hline
 Diabetes & 442 & 10 \\
 Compas & 6172 & 13 \\
 Seeds & 210 & 7 \\
 HTRU2 & 17898 & 8 \\
 \hline  
 MNIST & 60000 & (28, 28) \\
 Oxford Flowers & 8189 & various \\
 CIFAR10 & 60000 & (32, 32) \\
 CHD2R & 732 & (256, 256) \\
 \hline\hline
\end{tabular}
\caption{\label{table:datasets}Datasets summary}
\end{table}

In Table \eqref{table:datasets}, we have the datasets description regarding the number of samples and the number of attributes for each of the aforementioned datasets. At the same time, the number of classes (for classification problems) or the distribution of the targets (for regression problems) are not presented, due to the fact that our proposed autoencoder method is fully unsupervised. \\

We end this subsection by mentioning the parameters which will be used in our experiments regarding the unsupervised regressor consisting of different types of oblique trees. More precisely, in the following subsections, \texttt{max$\_$features} will denote the number of features drawn from the training dataset that are used to train each estimator, and \texttt{max$\_$samples} the number of samples which are drawn with replacement from the same training dataset for each base estimator. Additionally, \texttt{max$\_$depth} will denote the maximum depth of each tree (estimator), and \texttt{n$\_$estimators} will constitute the number of estimators (represented as oblique trees) which forms the entire ensemble. Also, for the datasets where we utilize the train-test splitting procedure, \texttt{test$\_$size} represents the size percentage of the test dataset, while \texttt{n$\_$train} and \texttt{n$\_$test} will denote the explicit train and test number of samples if these integer values are given.

\subsection{Tabular Data Reconstruction}
\label{subsec:tdrecon}

\begin{figure}[hb!]
\caption{Features reconstruction for the \textit{Diabetes} dataset}
\hspace{-1cm}
\begin{tabular}{cc}
    \raisebox{-.48\height}{\includegraphics[width=1.1\linewidth]{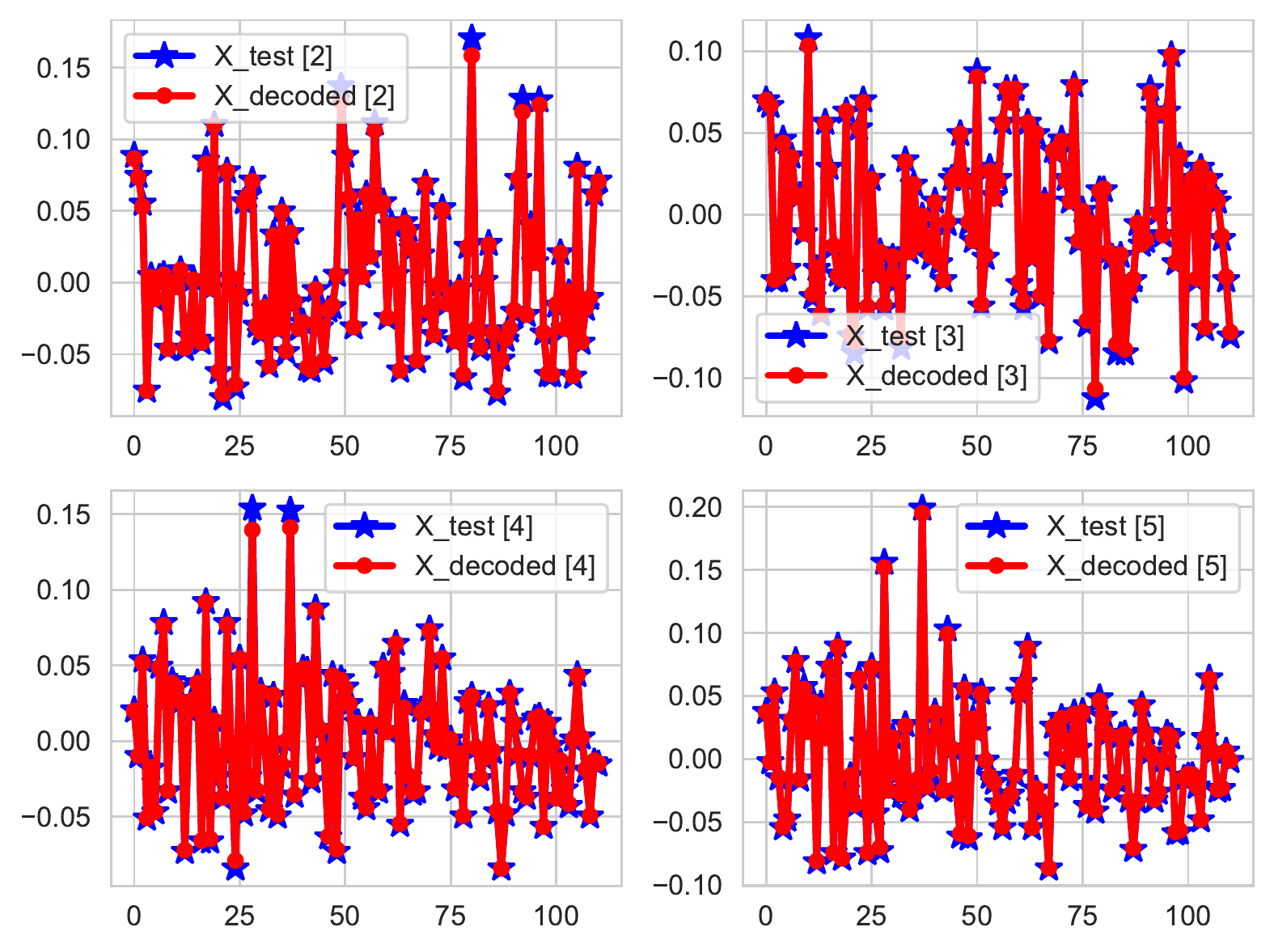}}
\end{tabular}
\label{fig:tabular_diabetes}
\end{figure}

In the present subsection we shall consider some basic experiments for the reconstruction of features belonging to various tabular datasets. Our first experiment is related to the \textit{Diabetes} dataset which consists of only $442$ number of samples. For this, we have set the test size percentage split to $0.25$, the number of estimators of the unsupervised oblique regressor to $200$, the maximum number of features percentage to $0.75$ and the number of bootstrap samples to $0.5$. Along with these, the maximum depth of each estimator is set to $3$. From figure \eqref{fig:tabular_diabetes} one observes that we obtain almost perfect reconstruction for some chosen numerical features.

Even though we have chosen a sizeable number of estimators, even with a maximum depth of $3$, the reconstruction of the real-valued features is performed with ease. \\

\begin{figure}[ht!]
\caption{Features reconstruction for the \textit{Compas} dataset}
\hspace{-0.5cm}
\begin{tabular}{cc}
    \raisebox{-.48\height}{\includegraphics[width=1.1\linewidth]{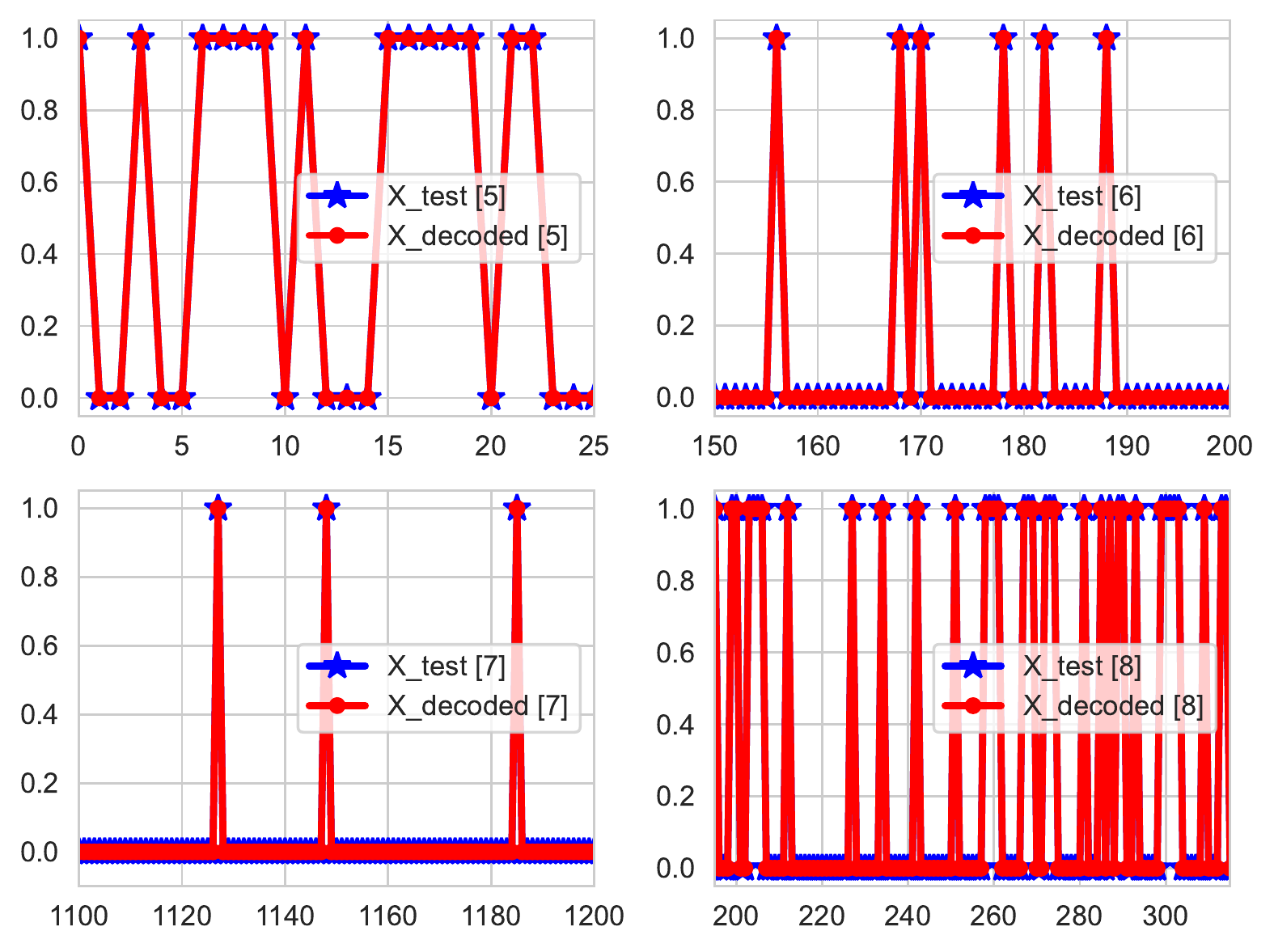}}
\end{tabular}
\label{fig:tabular_compas}
\end{figure}

For our second experiment, we have chosen the reconstruction of some categorical features on the \textit{Compas} dataset. For our simulation, we have considered the same parameters as in the previous experiment, with the exception that the train and test datasets were given indepdently. The results presented in figure \eqref{fig:tabular_compas} reveals the fact that even categorical features with various number of categories can be reconstructed with the appropriate number of estimators, even though the maximum depth of each estimator is set to a low value, namely $3$. 

It is worth mentioning that in the previously discussed experiments, we have used the \textit{HHCART} method for the oblique trees. For the \textit{Compas} dataset, we have utilized the \textit{PCA} method with number of components equal to $1$, while for the feature transformation for the \textit{HHCART} method on the \textit{Diabetes} dataset we have employed the \textit{Gaussian Random Projection} also with number of components equal to $1$. \\

\begin{figure*}[ht!]
\caption{Ablation study for the parameters of the \textbf{\textit{OF-AE}} method on the \textit{Diabetes} dataset}
\centering
\begin{tabular}{cc}
    \includegraphics[width=.32\linewidth]{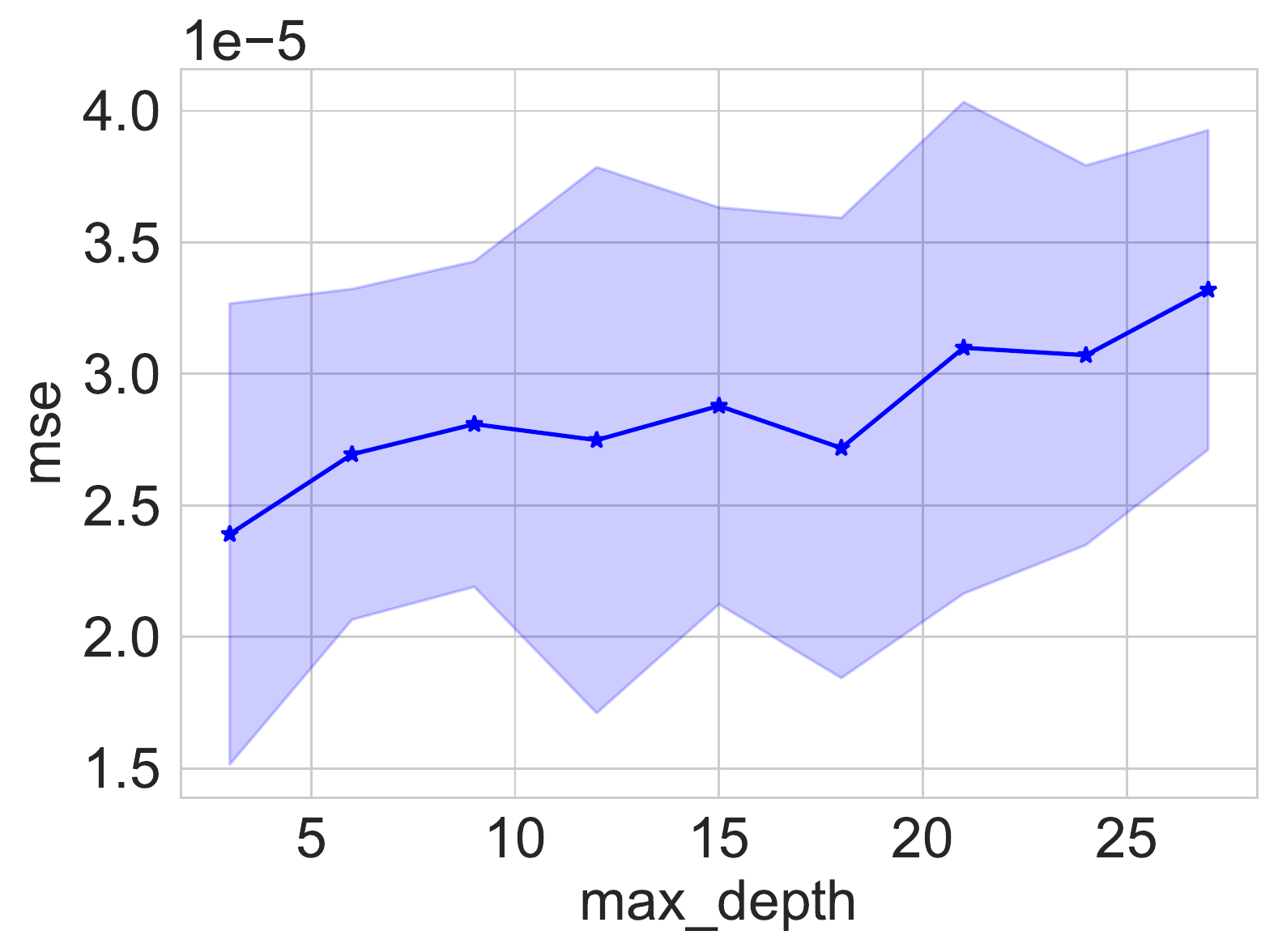} & \includegraphics[width=.32\linewidth]{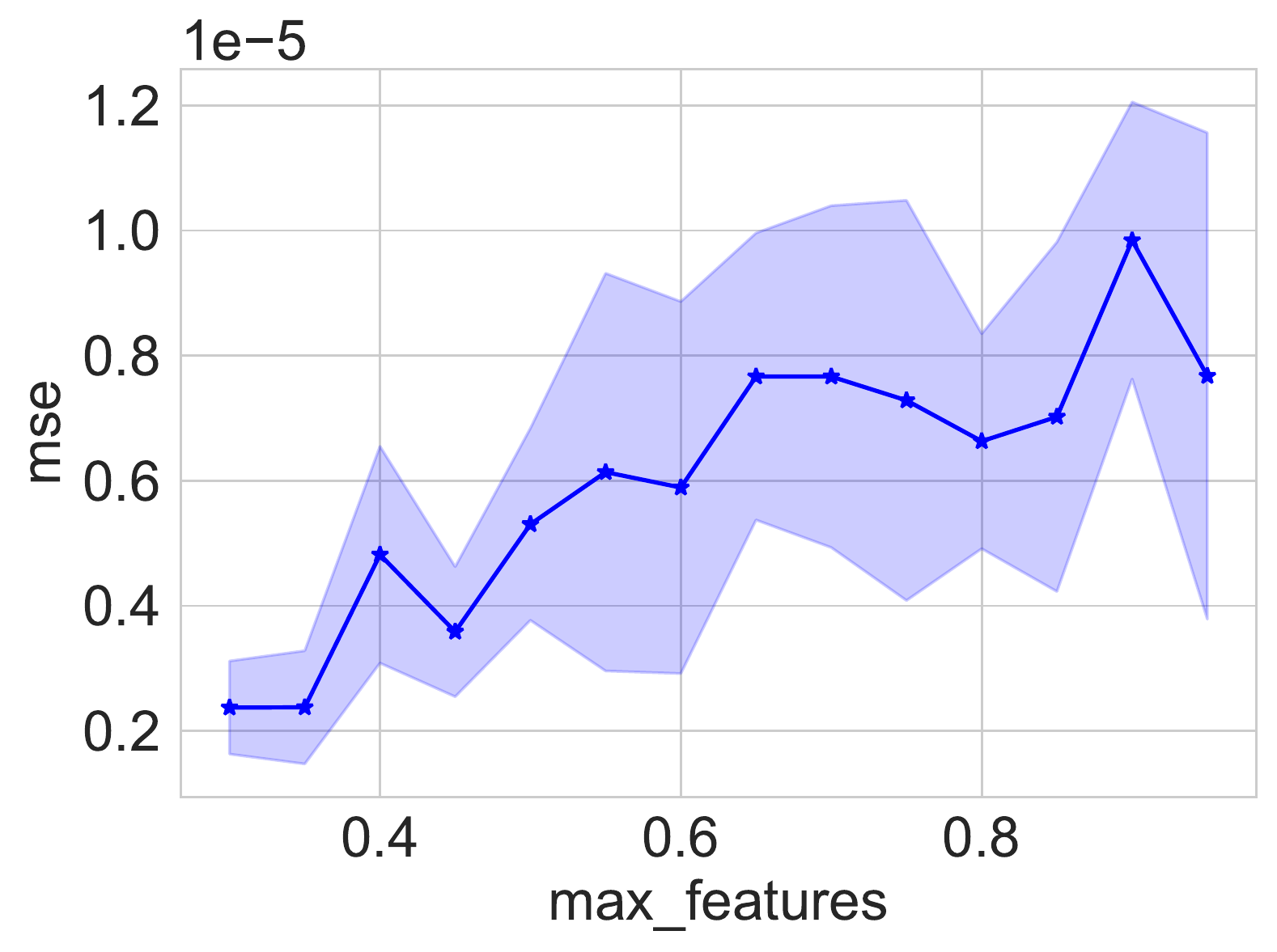} \\
        \includegraphics[width=.32\linewidth]{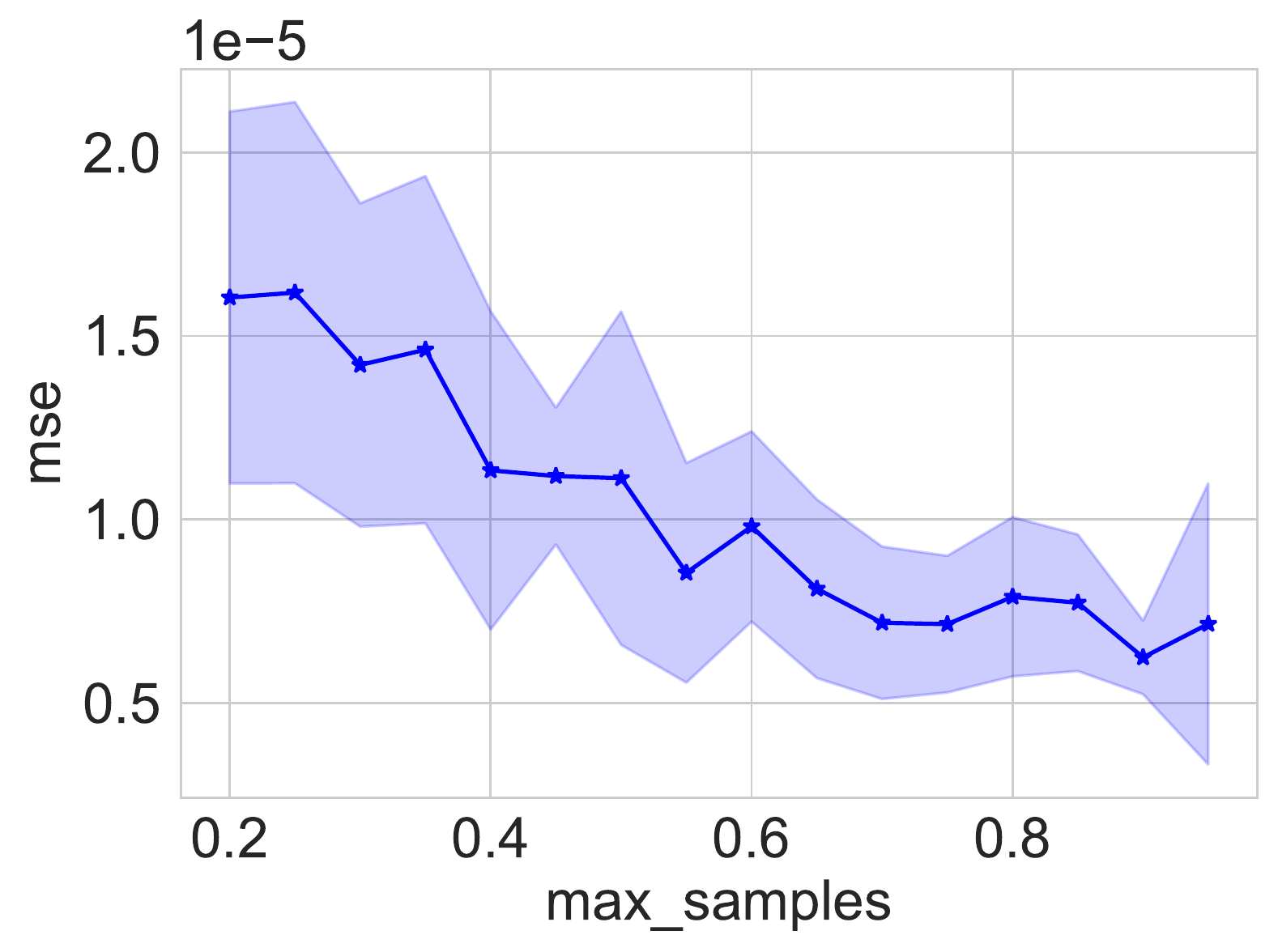} & \includegraphics[width=.32\linewidth]{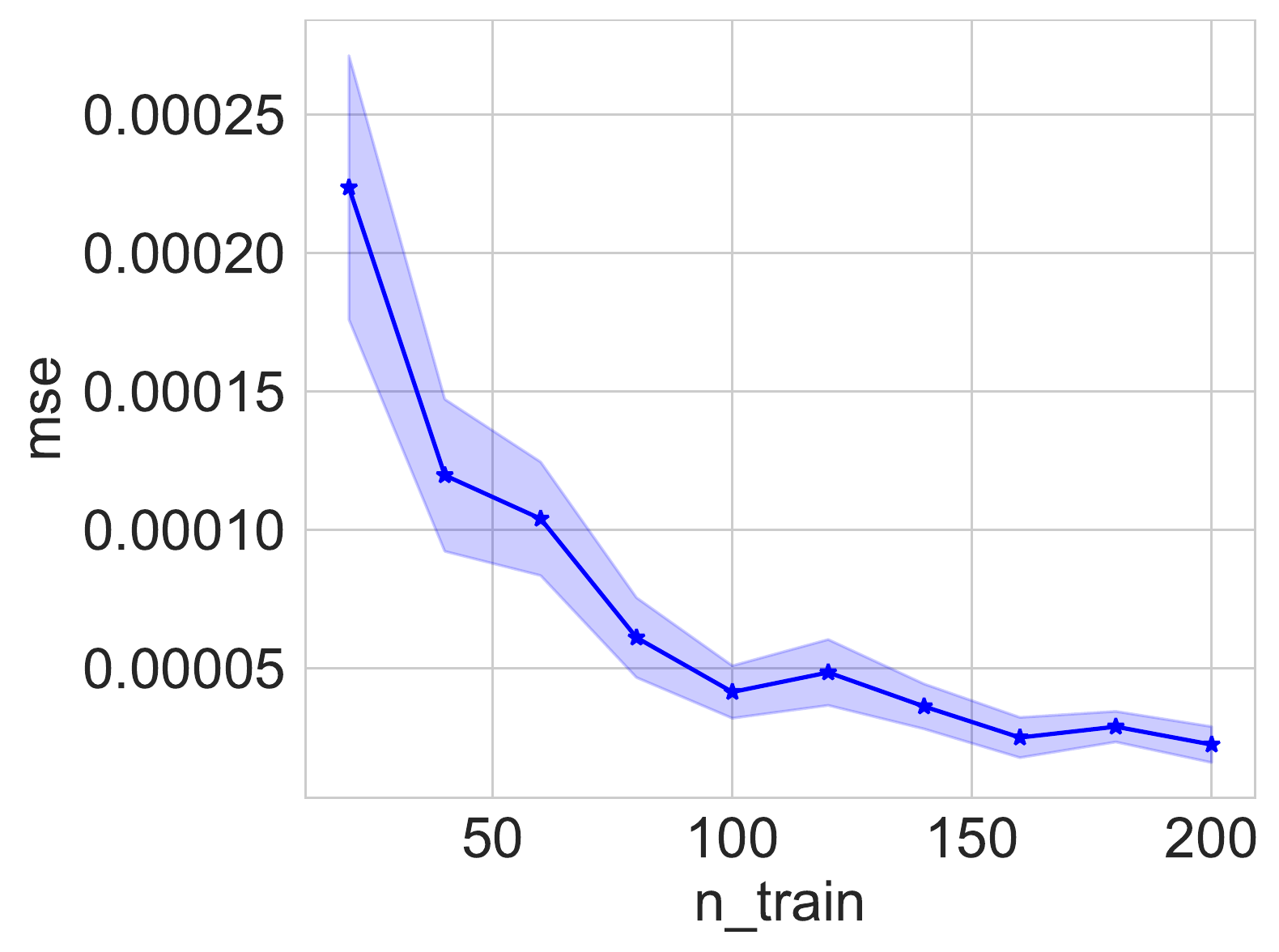}
\end{tabular}
\label{fig:tabular_ablation}
\end{figure*}

Now we will present some additional results for tabular datasets which will consist in the analysis of the impact of various parameters of the oblique trees ensemble. For this, let's consider the figure \eqref{fig:tabular_ablation}, where we have the variation of the \texttt{max$\_$depth}, \texttt{max$\_$features}, \texttt{max$\_$samples} and \texttt{n$\_$train} parameters with respect to the \textit{MSE} values of the decoded test samples reconstruction on the \textit{Diabetes} dataset. Even though we have not chosen the best values that will give us the lowest \textit{MSE}, figure \eqref{fig:tabular_ablation} faithfully highlights the importance of the parameters underpinning these methods. For the plot concerning the \texttt{max$\_$depth} parameter, we have chosen (\texttt{n$\_$estimators}, \texttt{ max$\_$samples}, \texttt{ max$\_$features}) = $(50, 0.5, 0.75)$, while for the \texttt{max$\_$features} plot, we have taken (\texttt{n$\_$estimators}, \texttt{ max$\_$samples}, \texttt{ max$\_$depth}) = $(100, 0.5, 3)$. 

At the same time, for the \texttt{max$\_$samples} plot we have set (\texttt{n$\_$estimators}, \texttt{ max$\_$features}, \texttt{ max$\_$depth}) = $(100, 0.75, 3)$, while for the \texttt{n$\_$train} plot, we have considered the fixed parameters (\texttt{n$\_$estimators}, \texttt{ max$\_$samples}, \texttt{ max$\_$features}, \texttt{ max$\_$depth}) = $(50, 0.5, 0.75, 3)$.
Also, for each plot we have made $10$ simulation runs (in order to obtain the mean and the standard deviation of our results), the \texttt{test$\_$size} was set to $0.25$, and the \textit{Gaussian Random Projection} was employed for the \textit{HHCART} oblique trees. \\
From figure \eqref{fig:tabular_ablation} one can easily see that the maximum depth has little influence on the reconstruction error in the situation when the number of estimators is set to a moderate value of $50$. Moreover, the maximum number of features seems to slightly increase the \textit{MSE} values. An explanation for this may be that a large percentage of the maximum features used reduces the inherent randomization of the ensemble methods, but one can't exclude the possibility that these results may appear due to the fact that the number of simulation runs is not large enough. On the other hand, the most decisive parameters are \texttt{max$\_$samples} and \texttt{n$\_$train}, respectively. More exactly, the \textit{MSE} values are monotonically decreasing when the maximum number of samples or the number of training data points are increased. This crucial observation will be further utilised in the design of the experiments regarding the reconstruction of images. \\

\begin{figure}[h!]
\caption{Results for fit $\&$ decode time for different oblique methods}
\centering
\includegraphics[width=0.4\textwidth]{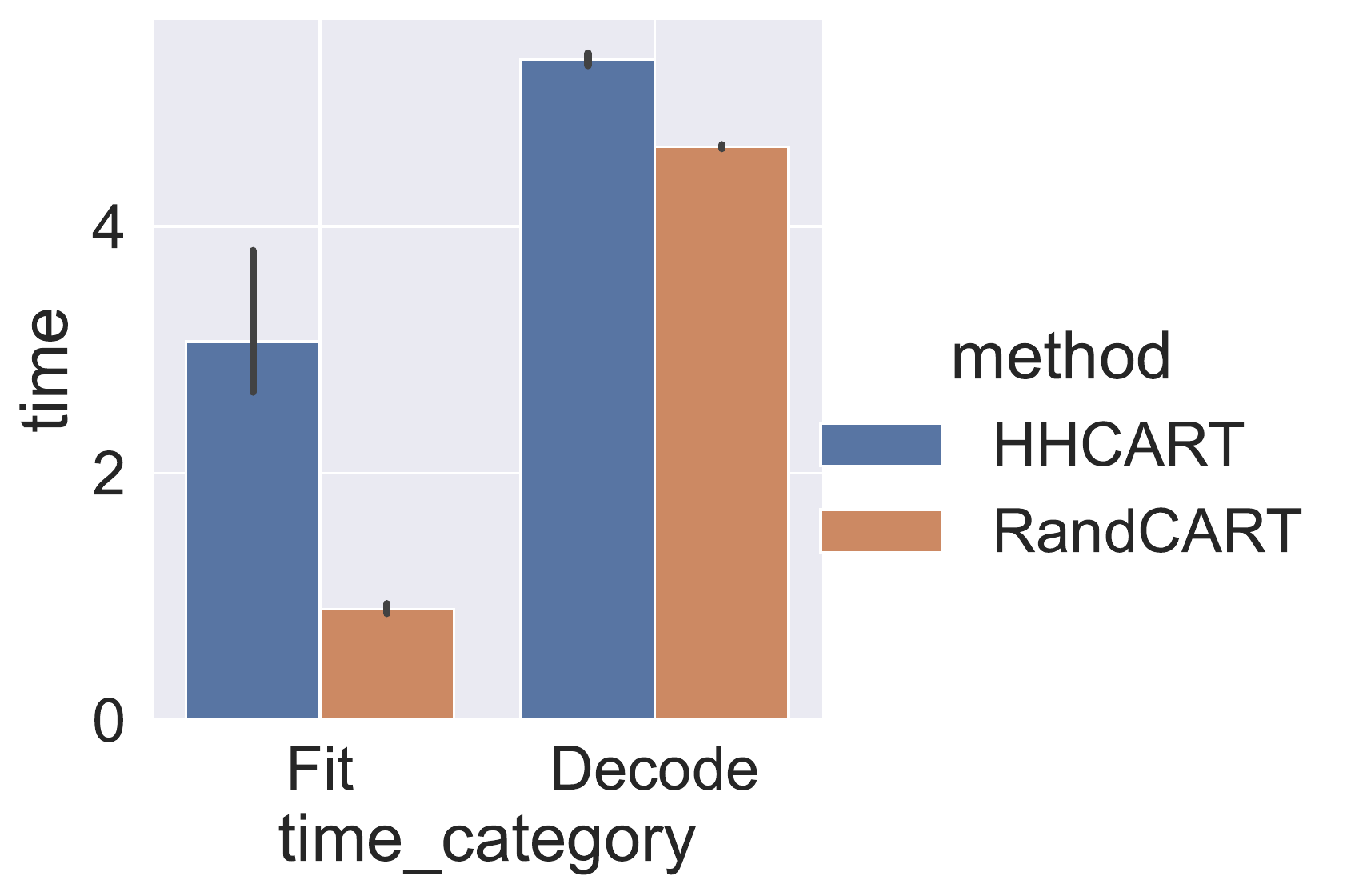}
\label{fig:tabular_barplot_time_seeds}
\end{figure}

For our next tabular data experiment, we have used the \textit{Seeds} dataset in order to compare the train and decode time values of the \textit{HHCART} and \textit{RandCART} oblique-type trees, respectively. Here, \texttt{n$\_$estimators} are set to $200$, \texttt{max$\_$samples} to $0.5$, \texttt{max$\_$features} to $0.75$ and \texttt{max$\_$depth} to $3$. For the \textit{Seeds} dataset which consists only in $210$ samples (where $25\%$ of them were used for the test data by setting the \texttt{test$\_$size} to $0.25$) the high number of trees, namely $200$, is utilized in order to emphasize the comparison of the running time of different types of oblique trees. In figure \eqref{fig:tabular_barplot_time_seeds} the line of the categories presented in the barplots represent the variation of $10$ simulation runs. From this plot, we can observe that the \textit{HHCART} method is much slower than the \emph{RandCART} technique at training and also at decoding. Furthermore, the major difference lies in the training time where there exists a large variation for the running time of the \textit{HHCART} method. 

\begin{figure}[ht!]
\caption{\emph{MSE} results with respect to the number of estimators}
\centering
\includegraphics[width=0.48\textwidth]{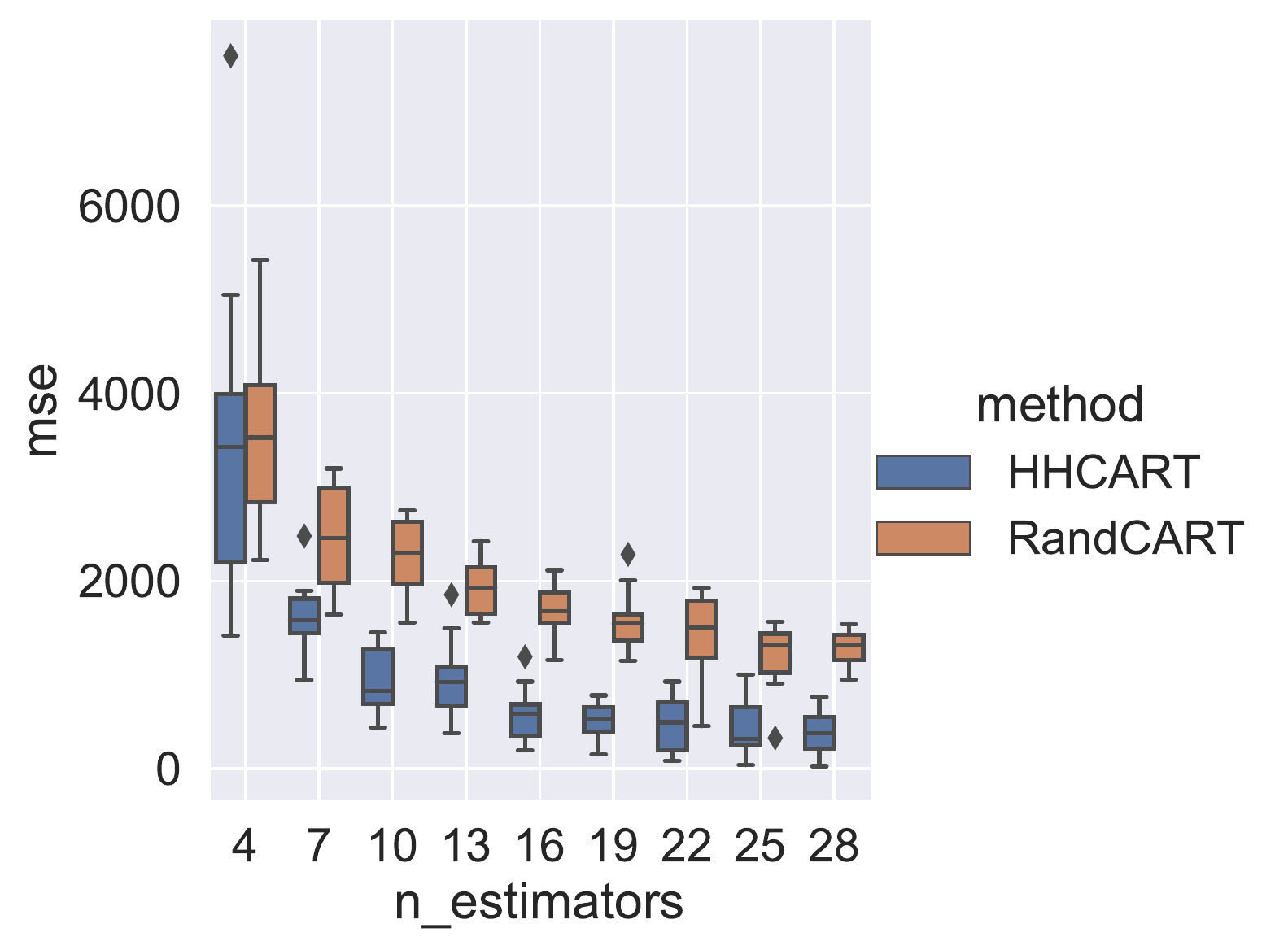}
\label{fig:tabular_boxplot_estimators_htru2}
\end{figure}

As we have seen, there are some noticeable differences between the fit time of the \textit{HHCART} method in comparison with the training time of the \textit{RandCART} method. Despite this fact, we are interested not only in how fast our ensemble methods are, but how precise they are with respect to reconstruction errors. We will end this subsection by investigating the differences in the \textit{MSE} values for the aforementioned oblique-type methods. In figure \eqref{fig:tabular_boxplot_estimators_htru2} we have used the \textit{HTRU2} dataset in order to compare the reconstruction values represented by the \textit{MSE}. Here, we have considered again $10$ simulation runs, $50$ sample points for training, $10$ test sample points for decoding, along with the following choices: (\texttt{max$\_$samples}, \texttt{ max$\_$features}, \texttt{ max$\_$depth}) = $(0.5, 0.75, 3)$. Despite the fact that the \textit{MSE} values are quite large (since we have chosen the previously mentioned parameters in order to observe the qualitative behaviour of our methods), the boxplots presented in figure \eqref{fig:tabular_boxplot_estimators_htru2} clearly show that the \textit{HHCART} method using the \textit{PCA} feature transformation gives better results (which means lower \textit{MSE} values) than \textit{RandCART} (it is worth mentioning that the reconstruction results for our autoencoder method are different than the prediction results for the same type of estimators as given in \cite{EODT}). But, in what we will see in the next subsection, we will use both of these methods for further comparisons.

\subsection{Image Reconstruction}
\label{subsec:imgrecon}

In this subsection, we shall apply a similar heuristic analysis  to the case of image reconstruction. For such situations, we will be able to precisely visualize the results. As for the case of the \emph{eForest} encoder \cite{eForest}, the \textbf{\textit{OF-AE}} method is suitable also for the reconstruction of images belonging to various datasets. In what follows, for the image reconstruction plots, the first row will represent some test samples, while in the second row we will always plot the reconstructed versions using \textbf{\textit{OF-AE}}.\\

For our first experiment, we consider the \textit{MNIST} dataset, where we have chosen the number of estimators equal to $300$, number of training samples $100$, and (\texttt{max$\_$samples}, \texttt{ max$\_$features}, \texttt{max$\_$depth}) = $(1.0, 0.75, 3)$. After the training, we have applied the usual encoding-decoding technique of \textbf{\textit{OF-AE}} on $10$ test samples. 

\begin{figure}[ht!]
\caption{Image reconstruction of \textit{MNIST} images}
\centering
\includegraphics[width=0.4\textwidth]{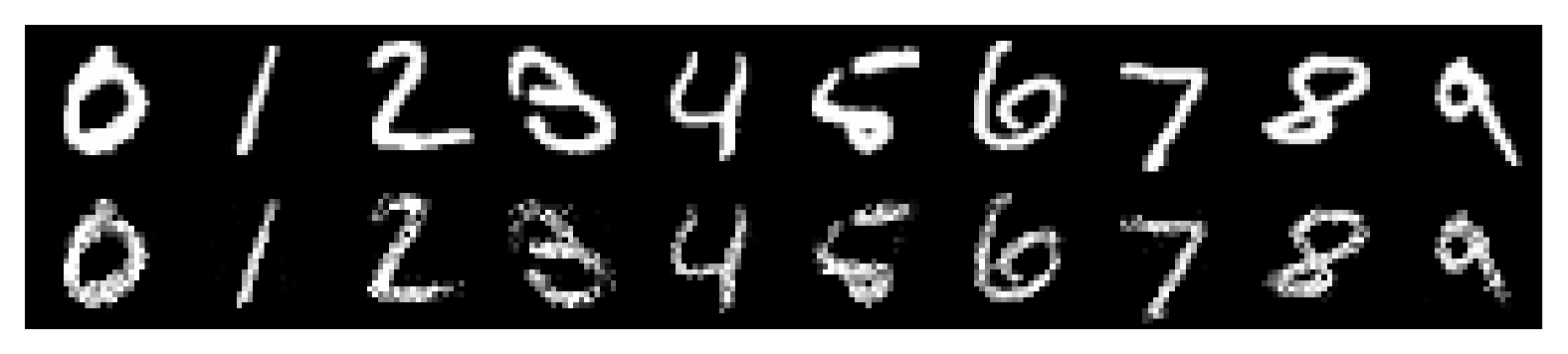}
\label{fig:mnist}
\end{figure}

From the depiction of figure \eqref{fig:mnist} we observe that we obtain a good representation of the test images even though we have used only $100$ training samples. A similar representation on the \textit{MNIST} dataset can be seen also in \cite{eForest} for the eForest encoder. The main difference is that we have used only $300$ estimators with a low maximum depth of $3$, both values being much lower than the ones from the \emph{eForest} encoder experiments. Even though for a better decoding quality we can increase the number of ensemble trees, our results suggest that the oblique splits can retain a compact representation of the features due to the inherent linear combination of features at each node split. \\

\begin{figure}[hb!]
\caption{Results of \textit{SSIM} values for different \textit{HHCART} feature space methods}
\centering
\includegraphics[width=0.32\textwidth]{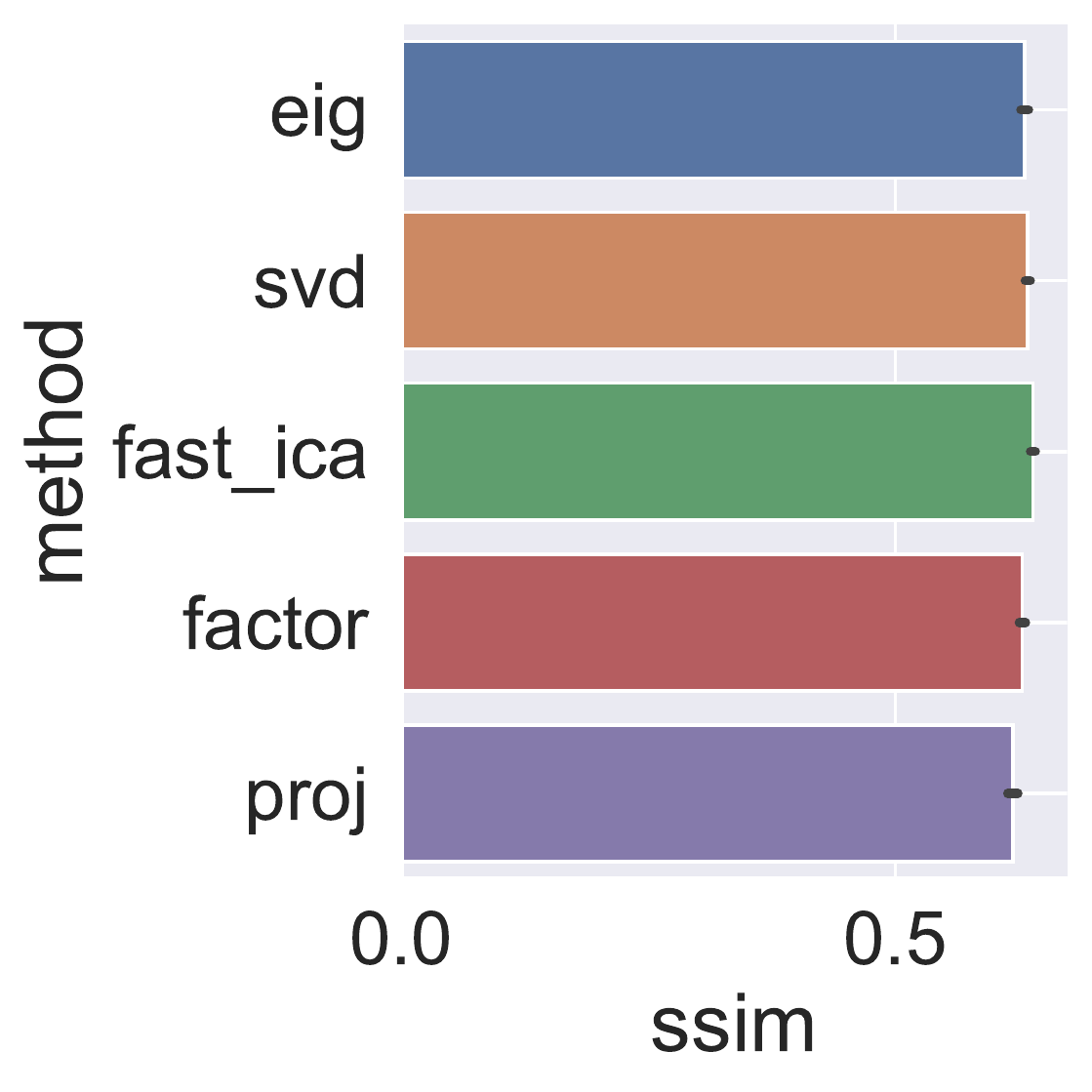}
\label{fig:images_barplot_ssim_eig_methods}
\end{figure}

Although in the previously mentioned image reconstruction experiment we have utilized the \textit{PCA} method for the feature transformation, in our second simulation we have compared the \emph{SSIM} values (again on the \emph{MNIST} dataset) for different types of methods for transforming the original space to the feature space. In \eqref{fig:images_barplot_ssim_eig_methods} we observe the \textit{SSIM} values of \textit{PCA} (denoted as \textit{eig}), alongside various implementations from \texttt{SKLearn}: \emph{Truncated-SVD} (briefly \textit{svd}), \emph{FastICA} (denoted as \textit{fast$\_$ica}), \textit{FactorAnalysis} (\textit{factor}) and \emph{Gaussian Random Projection} (denoted as \textit{proj}), respectively. In this simulation, \texttt{n$\_$estimators} and \texttt{n$\_$train} are set to $50$, while (\texttt{max$\_$samples}, \texttt{ max$\_$features}, \texttt{max$\_$depth}) = $(0.75, 0.75, 3)$.

What we easily observe is that the variation in the $10$ simulation runs we have experimented with is almost the same for every type of feature transformation method. Furthermore, it is noticeable the fact that the \emph{Gaussian Random Projection} seems to have the lowest impact on the reconstruction values. \\

\begin{figure}[h!]
\caption{\textit{SSIM} values with respect to the numbers of estimators}
\centering
\includegraphics[width=0.48\textwidth]{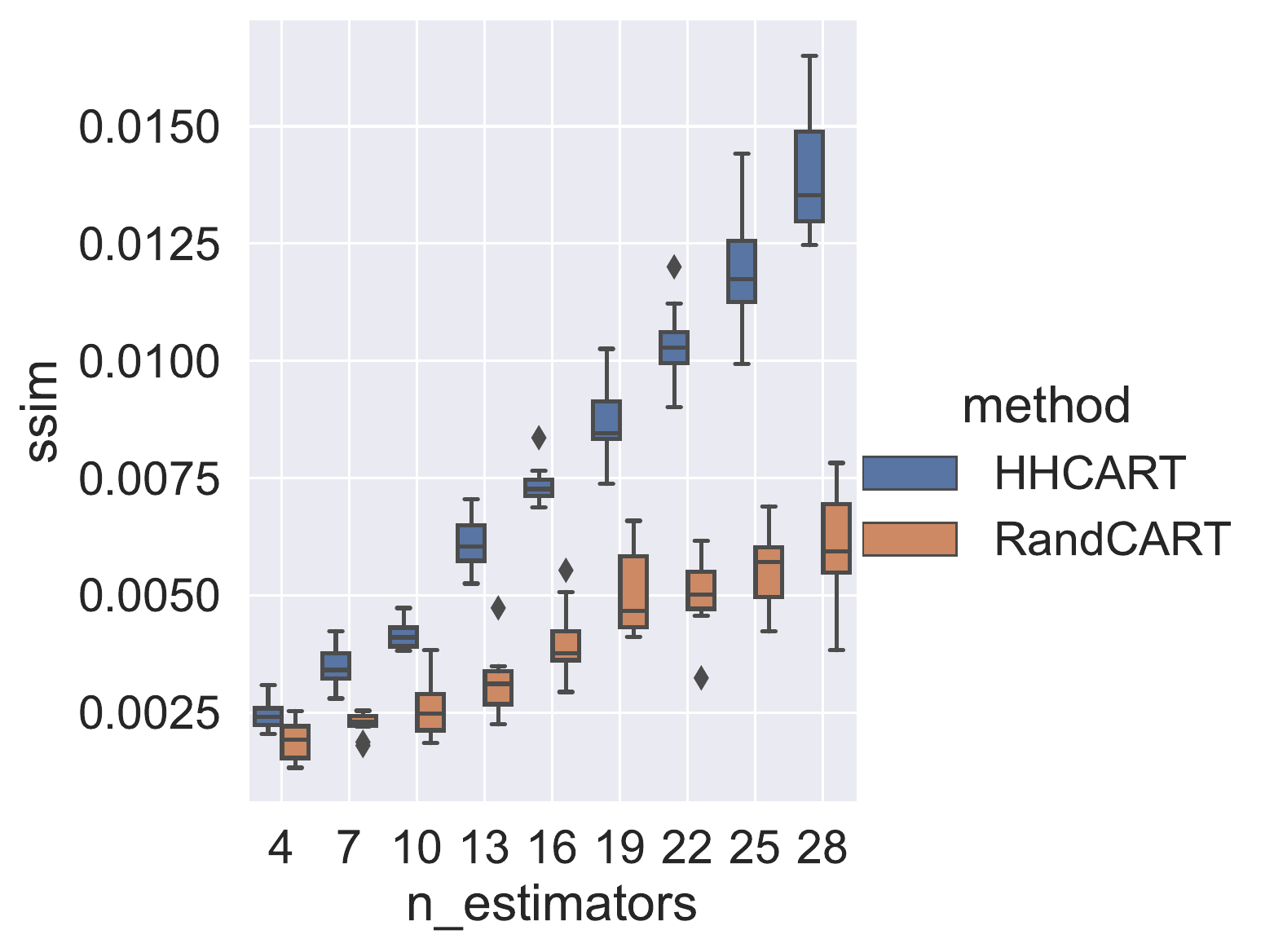}
\label{fig:images_boxplot_ssim_hhcart_vs_randcart}
\end{figure}

For our third experiment, we plan to evaluate the \emph{SSIM} values obtained from the decoded samples on the test dataset for the \emph{HHCART} (using the \emph{Gaussian Random Projection} method) and \emph{RandCART} oblique trees. In the experiment presented in figure \eqref{fig:images_boxplot_ssim_hhcart_vs_randcart} we have utilized our custom \emph{CHD2R} dataset where all the images are resized to $28 \times 28$. As in the case of the tabular data experiments, we took $10$ simulation runs and we fixed the parameters as (\texttt{max$\_$samples}, \texttt{ max$\_$features}, \texttt{max$\_$depth}) = $(0.25, 0.5, 30)$. If we compare the results of figure \eqref{fig:images_boxplot_ssim_hhcart_vs_randcart} with the ones from \eqref{fig:tabular_boxplot_estimators_htru2}, once more we observe the same behaviour: we get better results (this time in the case of \emph{SSIM} values) for the \emph{HHCART} method. Similar to the tabular dataset experiment, even though we can improve the results by increasing the number of estimators (or modifying other parameters), the results from figure \eqref{fig:images_boxplot_ssim_hhcart_vs_randcart} represent the main qualitative aspect of the difference between the aforementioned oblique methods.

\begin{figure}[ht!]
\caption{Training time of the \textbf{\textit{OF-AE}} methods}
\centering
\begin{tabular}{cc}
    \includegraphics[width=.58\linewidth]{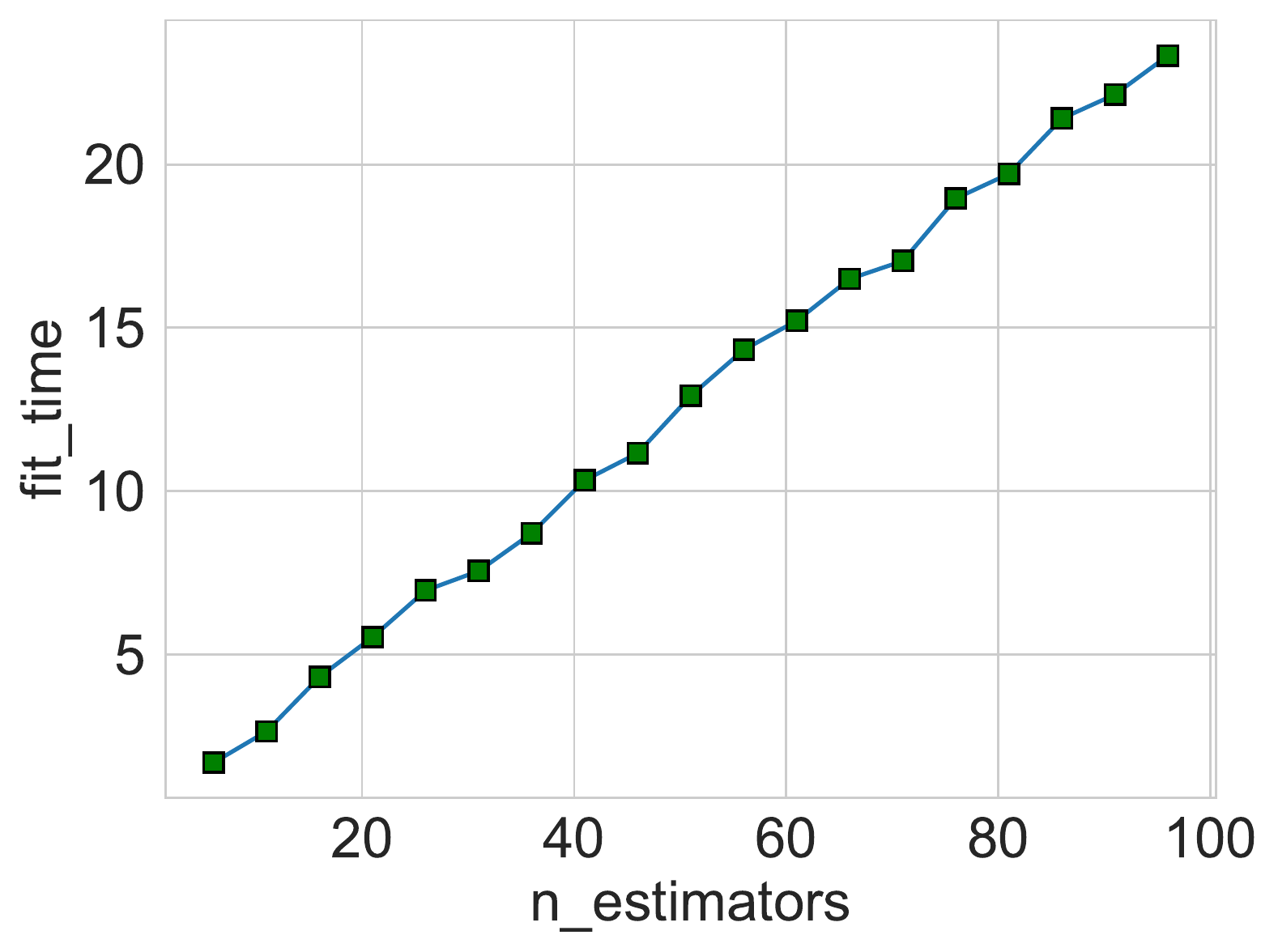} \\ \includegraphics[width=.58\linewidth]{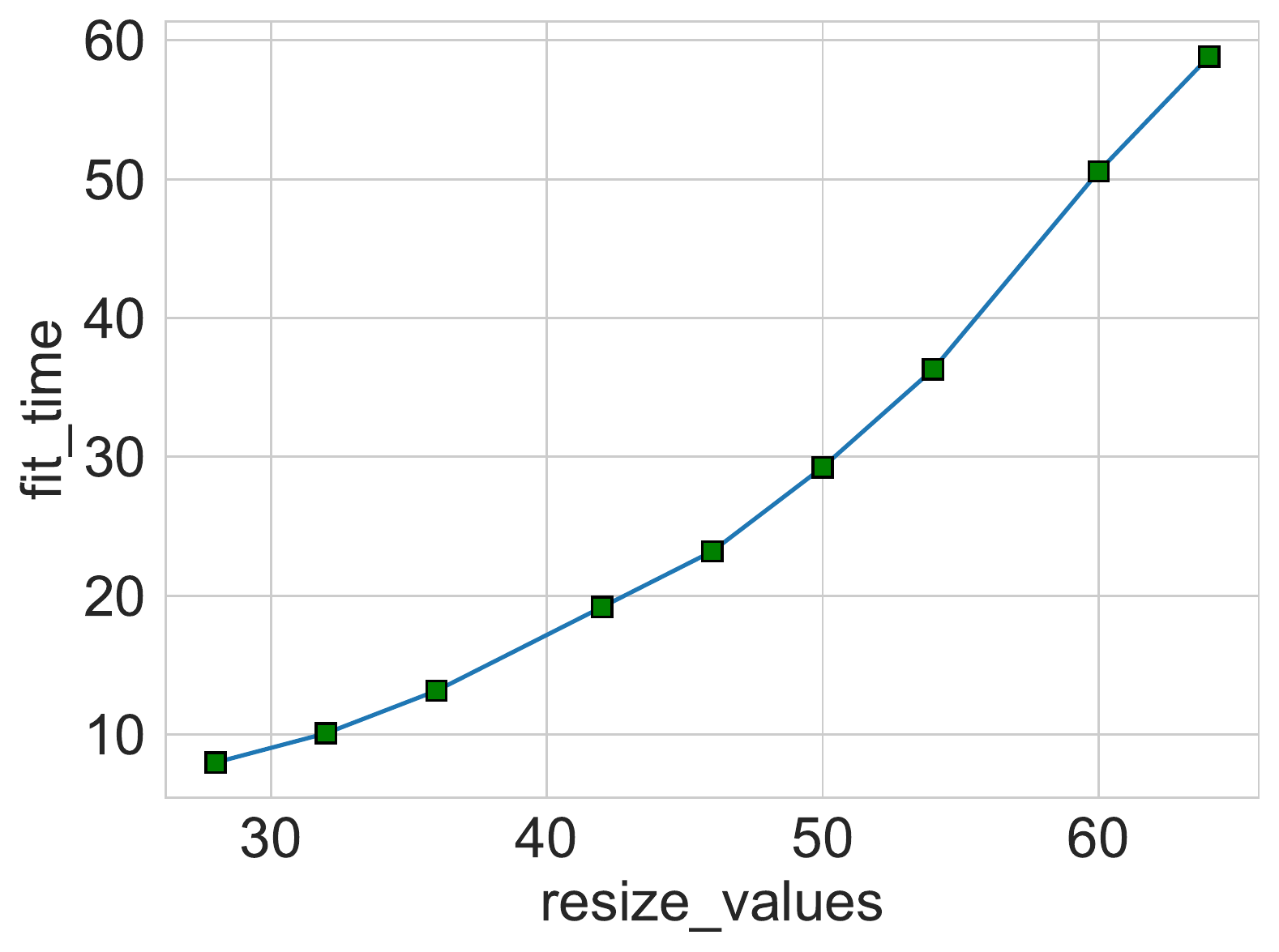}
\end{tabular}
\label{fig:ablation_images}
\end{figure}

Now, we will continue our ablation study with the investigation of the training time of \textbf{\textit{OF-AE}} method with respect to the number of estimators, along with the width and the height values of the images, respectively. For the case when we make the comparison of the image resize values versus the training time, we took (\texttt{n$\_$estimators}, \texttt{ max$\_$samples}, \texttt{ max$\_$features}, \texttt{max$\_$depth}) = $(30, 0.25, 1.0, 10)$. On the other hand, for the comparison between the number of estimators versus the training time, we chose (\texttt{resize$\_$value}, \texttt{max$\_$samples}, \texttt{ max$\_$features}, \texttt{max$\_$depth}) = $(28, 0.25, 1.0, 10)$.
For both situations, we have used the \emph{HHCART} oblique trees endowed with the \emph{Gaussian Random Projection} for transforming the original space to the feature space, along with $50$ number of training samples and $10$ number of test samples which needed to be decoded, on the \emph{Oxford Flowers} dataset. The results depicted in figure \eqref{fig:ablation_images} suggest that there exists a disadvantage our method, i.e. the training time grows in an accelerate manner if we increase the number of estimators or the width and height values of the images. The case when we increase the number of estimators is the most critical to us since the qualitative decoding representation of the autoencoder deeply depends on the number of oblique trees (this effectively represents that by increasing the number of estimators, we increase the number of constraints on the features through the weights equations). On the other hand, by increasing the size of the images, we observe that our method suffers from a "curse of dimensionality" problem. But, it is worth mentioning that the same phenomenon is also present for the \emph{eForest} encoder (where, for the same reconstruction quality of the decoded images, the number of axis-parallel trees is much larger) and for the tree-type encoder presented in \cite{AguilarAE} (where for each feature we must construct a classification decision tree with the target labels as possible pixel values).

\begin{figure*}[ht!]
\caption{Ablation study on \textit{CIFAR10} dataset}
\hspace{0.55cm}
\begin{tabular}{cc}
    \includegraphics[width=.46\linewidth]{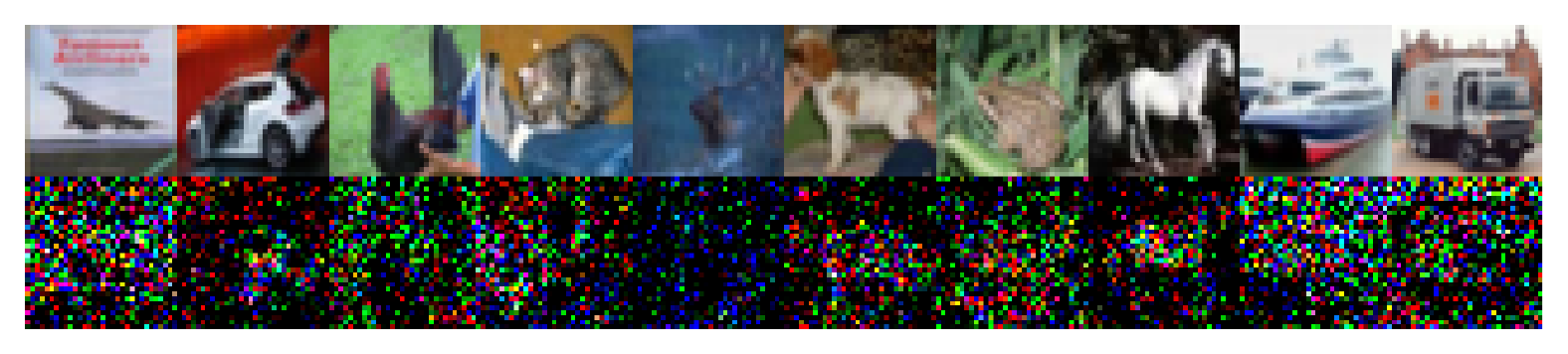} & 
    \includegraphics[width=.46\linewidth]{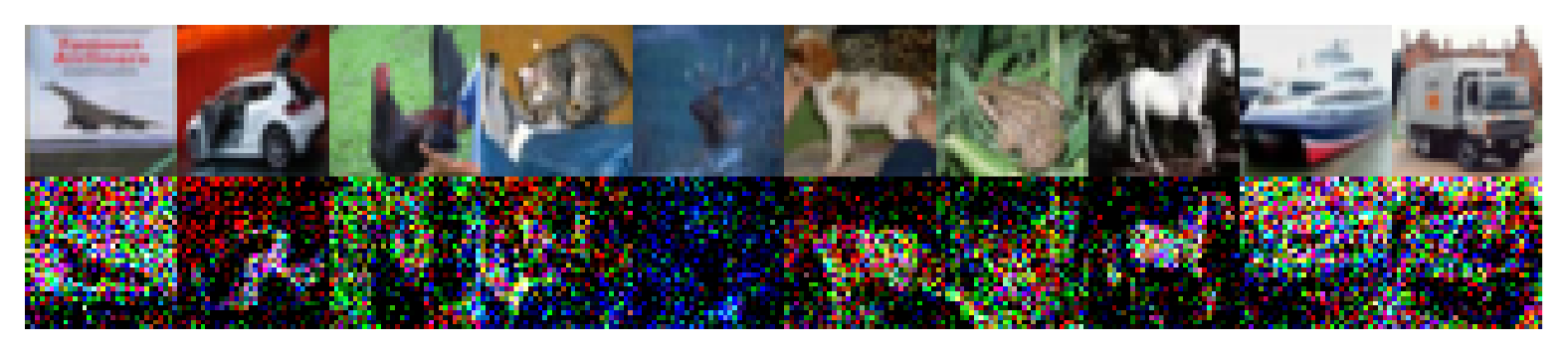} \\
    \includegraphics[width=.46\linewidth]{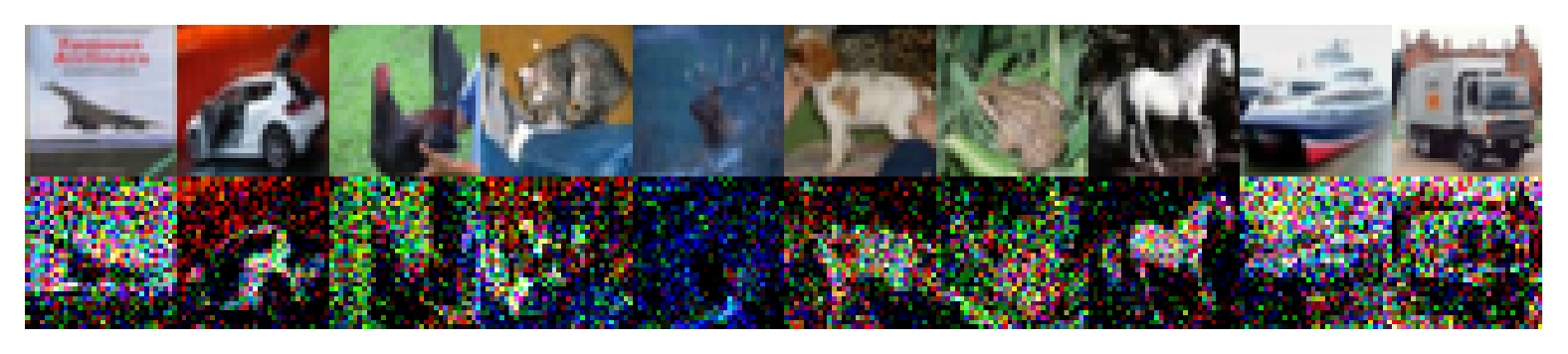} &
        \includegraphics[width=.46\linewidth]{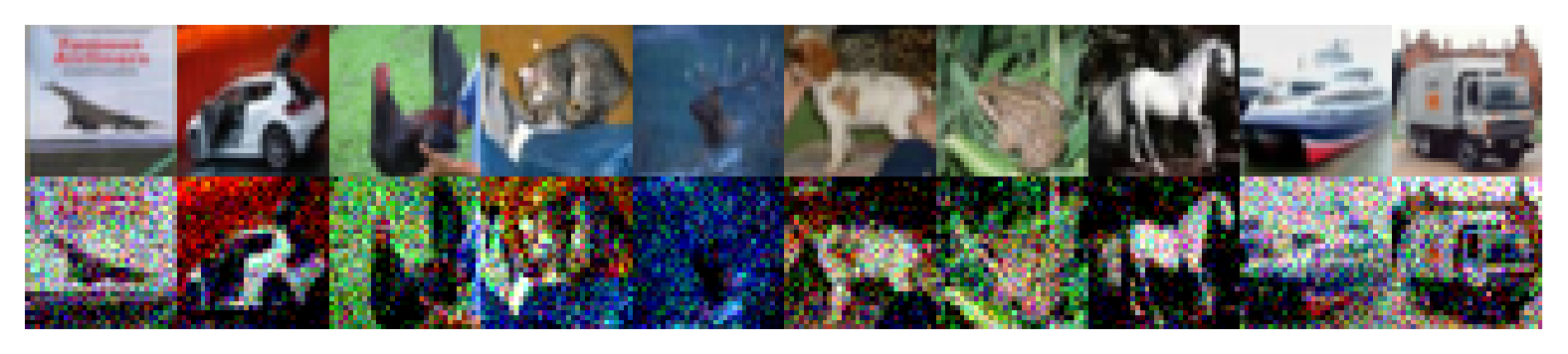} \\
        \includegraphics[width=.46\linewidth]{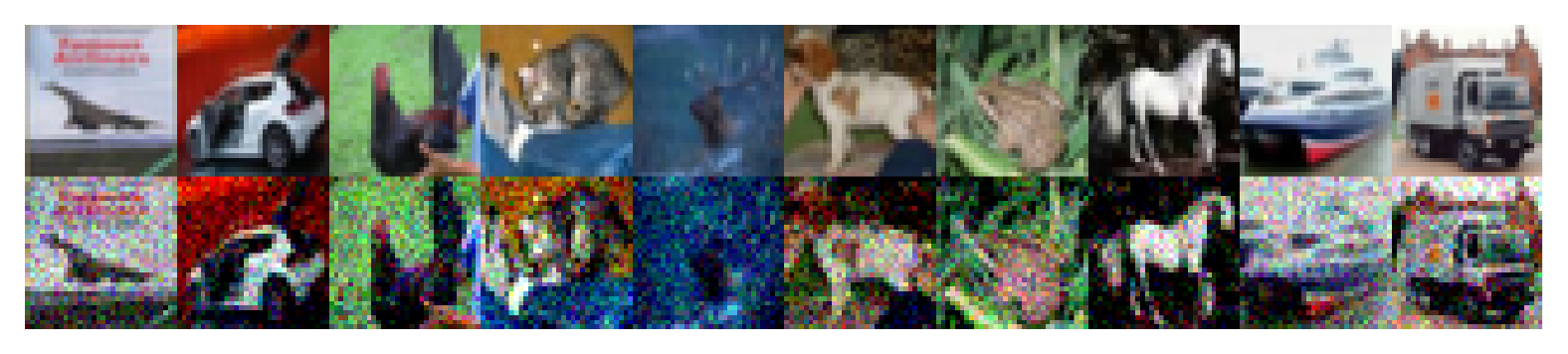} &
        \includegraphics[width=.46\linewidth]{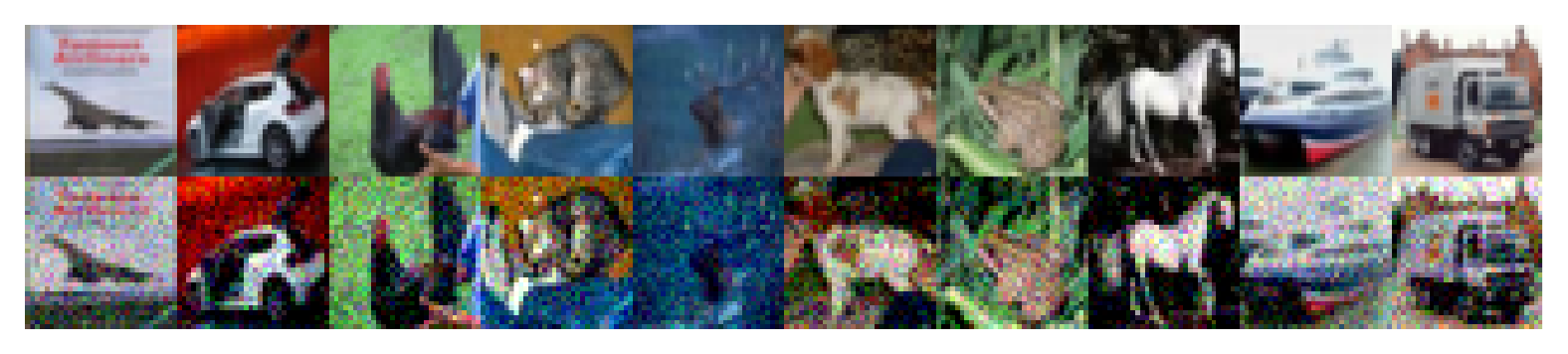}
\end{tabular}
\label{fig:cifar10_ablation}
\end{figure*}

For our second to last experiment, we will consider some simulations on the $32 \times 32$ images from the \emph{CIFAR10} dataset using ensemble of \emph{RandCART} oblique trees with maximum depth of $3$. This time we are clearly able to visualize the decoding of $10$ test samples with respect to various choices of the parameters of our \textbf{\textit{OF-AE}} method. In table \eqref{table:cifar10_parameters} we have $6$ experiment versions with different parameters on the \emph{CIFAR10} dataset. Furthermore, for these versions we have the decoded samples presented in figure \eqref{fig:cifar10_ablation}, where, from left to right, on the first row we have the results of experiment versions v3 and v1, for the second row we have the versions v2 and v4, while in the last row we eventually see the versions v5 and v6, respectively. \\
From the first row, it is easy to see that, for a low number of estimators, the results are unsatisfactory. 

In addition, for the transition between v3 and v1 we get qualitative decoding improvements by increasing the number of estimators.
On the other hand, let's look at the results from the second row. Here, we have the v2 and v4 experiment versions, where the transition between them is given by decreasing the number of training samples and increasing the number of estimators, which suggest that the number of estimators have a larger influence over the decoding process than the number of training samples. Finally, in the last row, the transition between v5 and v6 is given by increasing the number of estimators and the number of training samples which leads to even better results. But, we must mention that for increasing the quality of the reconstructed images, in this transition we have decreased also the \texttt{max$\_$samples} parameter.

\begin{table}[h!]
\normalsize
\resizebox{\columnwidth}{1.9cm}{%
\begin{tabular}{ ||p{1cm}||p{1cm}||p{1.7cm}||p{1.7cm}||p{1.7cm}||  }
 \hline\hline
 version & n$\_$train & n$\_$estimators & max$\_$samples & max$\_$features \\
 \hline\hline
 v1 & 300 & 300 & 0.5 & 0.25 \\
 v2 & 600 & 300 & 0.5 & 0.25 \\
 v3 & 1000 & 100 & 0.5 & 0.25 \\
 v4 & 200 & 800 & 0.75 & 0.25 \\
 v5 & 600 & 800 & 0.75 & 0.25 \\
 v6 & 800 & 1000 & 0.5 & 0.25 \\
 \hline\hline
\end{tabular}%
}
\caption{\label{table:cifar10_parameters}Parameters for the \emph{CIFAR10} dataset}
\end{table}

\begin{figure}[h!]
\caption{Model reuse on the \textit{Oxford Flowers} $\&$ \textit{CHD2R} datasets}
\centering
\begin{tabular}{c}
    \includegraphics[width=.99\linewidth]{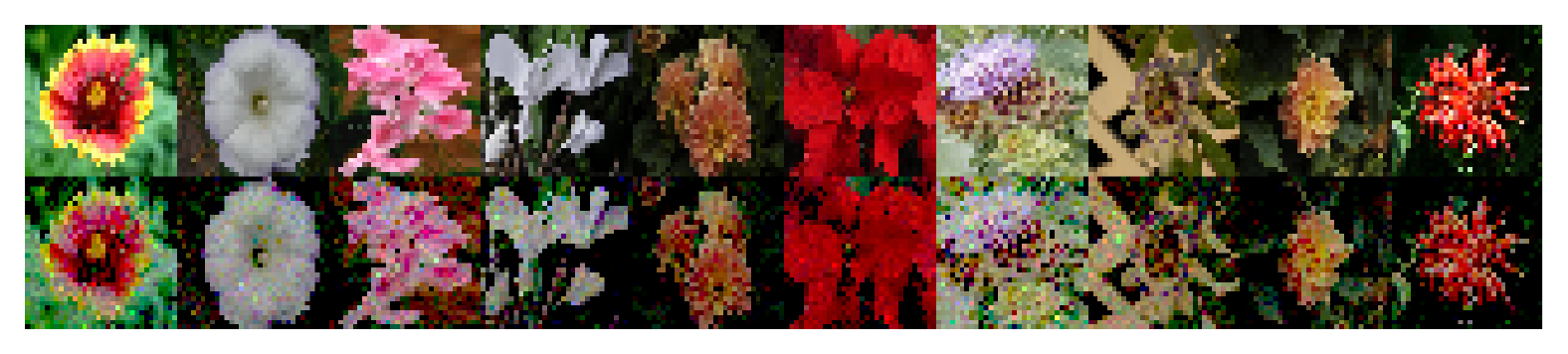} \\
    \includegraphics[width=.99\linewidth]{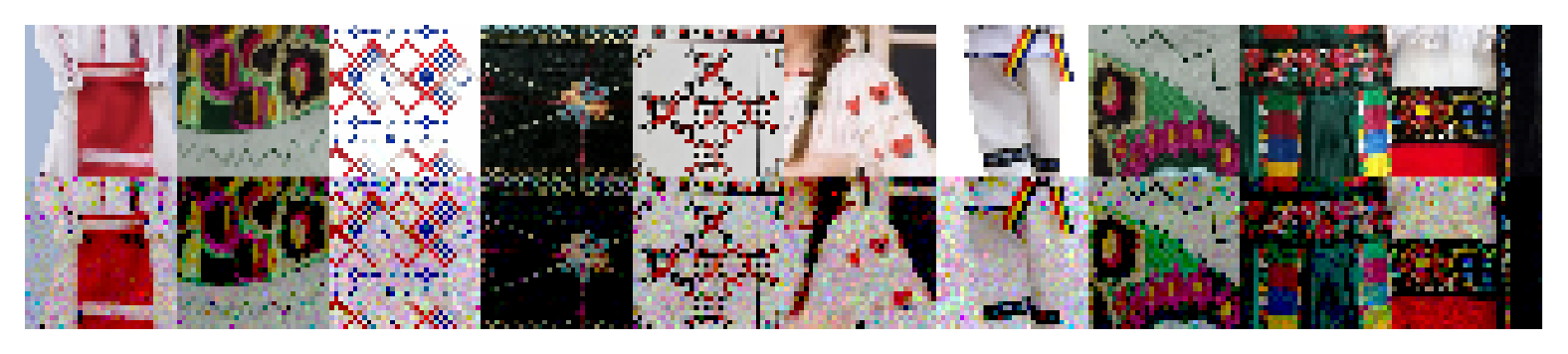}
\end{tabular}
\label{fig:model_reuse}
\end{figure}

Our last experiment is the same as the one from \cite{eForest} for the \emph{eForest} encoder, regarding model reuse. 
In our case, we have trained our \textbf{\textit{OF-AE}} method on $1000$ training samples belonging to the \emph{CIFAR10} dataset. After the training we have decoded $10$ test samples (resized to  the \emph{CIFAR10} images dimensions $32 \times 32$) belonging to the \emph{Oxford Flowers} and \emph{CHD2R} dataset, respectively. Moreover, by setting the parameters (\texttt{n$\_$estimators}, \texttt{ max$\_$samples}, \texttt{ max$\_$features}, \texttt{ max$\_$depth}) = $(1000, 0.5, 0.25, 3)$ and using the \emph{HHCART} method with the \emph{Gaussian Random Projection} feature transformation method, we get the results from figure \eqref{fig:model_reuse}. 
Finally, despite the fact that our results can be improved by increasing much more the number of estimators, as with the \emph{eForest} encoder, our \textbf{\textit{OF-AE}} method trained on a small number of images is able to reproduce almost exactly images pertaining to two totally different datasets.

\section{Conclusions and Discussions}
\label{sec:conc}

In this paper we have extended the \emph{eForest} encoder to unsupervised ensembles of oblique trees. By utilizing some particular type of multivariate trees, we have shown that it is possible to devise a forest-type autoencoder through constrained optimization problems. Our method \textbf{\textit{OF-AE}} has the same limitations as the \emph{eForest} encoder. More precisely, as we have shown in the previous section, the training time depends heavily on the image dimensions and on the number of estimators, thus our method becomes unfeasible for datasets consisting of images with large width and height. Furthermore, this problem appears also for other different tree-type autoencoders, i.e. like the one from \cite{AguilarAE} where for each feature one must construct a classification decision tree. The advantage of \textbf{\textit{OF-AE}} ensemble method is that it can be trained in an unsupervised manner, as in the case of the random forest-type embedding variant of the \emph{eForest} encoder. \\
Up to our knowledge, the method introduced in this paper is the only forest-type autoencoder where the encoding-decoding process is represented by a constrained optimization problem. Therefore, this article represents a significant departure from the works \cite{Bennett}, \cite{Bennett-Optimal} and \cite{Brown} (where the authors constructed the training process of the decision trees using optimization problems) since our algorithm requires the encoding-decoding procedure (and not the training) to be represented as an optimization problem. \\
Finally, it is worth investigating if our methodology can be extended to neural-type trees as in \cite{DNDF} and \cite{TannoANT}, respectively. In addition, a possibility would be to expand our work by drawing upon the methods used in \cite{TreeArgmin} where the discrete tree parameters are learned through an optimization process represented by a global loss function.



%
\bibliographystyle{IEEEtran}
\bibliography{references}

\clearpage

\end{document}